\pdfoutput=1

\documentclass[11pt]{article}

\usepackage[final]{acl}

\usepackage{times}
\usepackage{latexsym}

\usepackage[T1]{fontenc}

\usepackage[utf8]{inputenc}

\usepackage{microtype}

\usepackage{inconsolata}

\usepackage[hang,flushmargin]{footmisc}

\title{\uncle: Benchmarking Uncertainty Expressions \\ in Long-Form Generation}

\author{
Ruihan Yang\textsuperscript{1}\thanks{Equal contribution, listed in alphabetical order. Work done during Tencent AI Lab internship. \dag Corresponding authors.}, 
Caiqi Zhang\textsuperscript{2}\footnotemark[1], 
Zhisong Zhang\textsuperscript{4}\textsuperscript{\dag},
\textbf{Xinting Huang}\textsuperscript{3}, \\
\textbf{Dong Yu}\textsuperscript{3}, 
\textbf{Nigel Collier}\textsuperscript{2}\textsuperscript{\dag}, 
\textbf{Deqing Yang}\textsuperscript{1}\textsuperscript{\dag} \\
\textsuperscript{1}Fudan University \quad
\textsuperscript{2}University of Cambridge \quad \\
\textsuperscript{3}Tencent AI Lab \quad
\textsuperscript{4}City University of Hong Kong\\
{\small \texttt{\{rhyang17,yangdeqing\}@fudan.edu.cn$^{1}$, \{cz391,nhc30\}@cam.ac.uk$^{2}$, zhisong.zhang@cityu.edu.hk$^{4}$}}
}

\usepackage{microtype}
\usepackage{graphicx}
\usepackage{arydshln}
\usepackage{inconsolata}
\usepackage{todonotes}
\usepackage{tabularx}
\usepackage{booktabs} 
\usepackage{multirow}
\usepackage{amsmath}
\usepackage{amssymb}
\usepackage{graphicx}
\usepackage{subcaption}
\usepackage{fge}

\usepackage[framemethod=TikZ]{mdframed} 
\usepackage{listings}
\usepackage{longtable}
\usepackage{tcolorbox}
\usepackage{xspace}
\usepackage{wrapfig}
\usepackage{pgfplots}
\usepackage{color, colortbl}
\usepackage{array}
\usepackage{paralist}
\usepackage{geometry}
\usepackage{enumitem}
\usepackage{float}  
\usepackage{makecell}
\usepackage{pifont} 

\newcommand{\rparagraph}[1]{\vspace{1.2mm}\noindent\textbf{#1.}}

\definecolor{Gray}{gray}{0.92}
\definecolor{racing-green}{rgb}{0.0, 0.8, 0.6}
\definecolor{awesome-red}{rgb}{1.0, 0.13, 0.32}
\definecolor{LightCyan}{rgb}{0.88,1,1}
\definecolor{darkgreen}{RGB}{0,150,0}
\definecolor{Ground}{RGB}{255,184,55}
\definecolor{Dirt}{RGB}{191,169,115}
\definecolor{Pink}{RGB}{226,184,176}
\definecolor{Violet}{RGB}{163,148,170}
\definecolor{darkred}{RGB}{150,0,0} 
\definecolor{Red}{RGB}{171, 61, 56}
\definecolor{Green}{RGB}{62, 139, 117}
\definecolor{Blue}{RGB}{48, 110, 184}

\definecolor{CC}{RGB}{198, 226, 212} 
\definecolor{UU}{RGB}{198, 228, 253} 
\definecolor{CU}{RGB}{247, 202, 193} 
\definecolor{UC}{RGB}{242, 224, 253}

\newcommand{\ie}{\textit{i}.\textit{e}.,\ }
\newcommand{\eg}{\textit{e}.\textit{g}.,\ }

\newtheorem{metric}{Metric}

\definecolor{level4}{RGB}{110,136,203}
\definecolor{level3}{RGB}{173,190,226}
\definecolor{level2}{RGB}{205,208,243}
\definecolor{level1}{RGB}{236,236,252}

\newcommand{\cellcolorval}[1]{
   \ifdim#1pt>100pt \cellcolor{level4!95}#1\relax
   \else
   \ifdim#1pt>90pt \cellcolor{level4!85}#1\relax
   \else\ifdim#1pt>80pt \cellcolor{level3!75}#1\relax
   \else\ifdim#1pt>70pt \cellcolor{level3!60}#1\relax
   \else\ifdim#1pt>60pt \cellcolor{level2!70}#1\relax
   \else\ifdim#1pt>50pt \cellcolor{level2!45}#1\relax
   \else\ifdim#1pt>40pt \cellcolor{level2!30}#1\relax
   \else\ifdim#1pt>30pt \cellcolor{level2!10}#1\relax
   \else\ifdim#1pt>20pt \cellcolor{level1!10}#1\relax
   \else \cellcolor{level1!0}#1\relax
   \fi\fi\fi\fi\fi\fi\fi\fi
}

\newcolumntype{g}{>{\columncolor{Ground!7}}c}
\newcolumntype{d}{>{\columncolor{cyan!6}}c}
\newcolumntype{f}{>{\columncolor{lime!6}}c}
\newcolumntype{v}{>{\columncolor{purple!6}}c}
\newcolumntype{u}{>{\cellcolorval}c}

\newcommand{\uncle}{\textsc{UncLE}\xspace}

\begin{document}
\maketitle
\begin{abstract}
Large Language Models (LLMs) are prone to hallucination, particularly in long-form generations. A promising direction to mitigate hallucination is to teach LLMs to express uncertainty explicitly when they lack sufficient knowledge. However, existing work lacks direct and fair evaluation of LLMs' ability to express uncertainty effectively in long-form generation. 
To address this gap, we first introduce \uncle, a benchmark designed to evaluate uncertainty expression in both long- and short-form question answering (QA). \uncle covers five domains and includes more than 1{,}000 entities, each with paired short- and long-form QA items. Our dataset is the first to directly link short- and long-form QA through aligned questions and gold-standard answers.
Along with \uncle, we propose a  suite of new metrics to assess the models' capabilities to selectively express uncertainty. We then demonstrate that current models fail to convey uncertainty appropriately in long-form generation. We further explore both prompt-based and training-based methods to improve models' performance, with the training-based methods yielding greater gains. Further analysis of alignment gaps between short- and long-form uncertainty expression highlights promising directions for future research using \uncle. \raisebox{-0.2\height}{\includegraphics[height=1.1em]{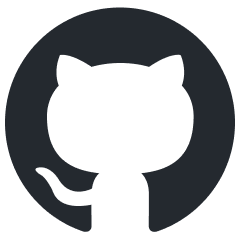}} 
[\href{https://github.com/rhyang2021/UNCLE}{Project Homepage}]

\end{abstract}

\section{Introduction}

\begin{figure}[t]
    \centering
    \includegraphics[width=0.97\columnwidth]{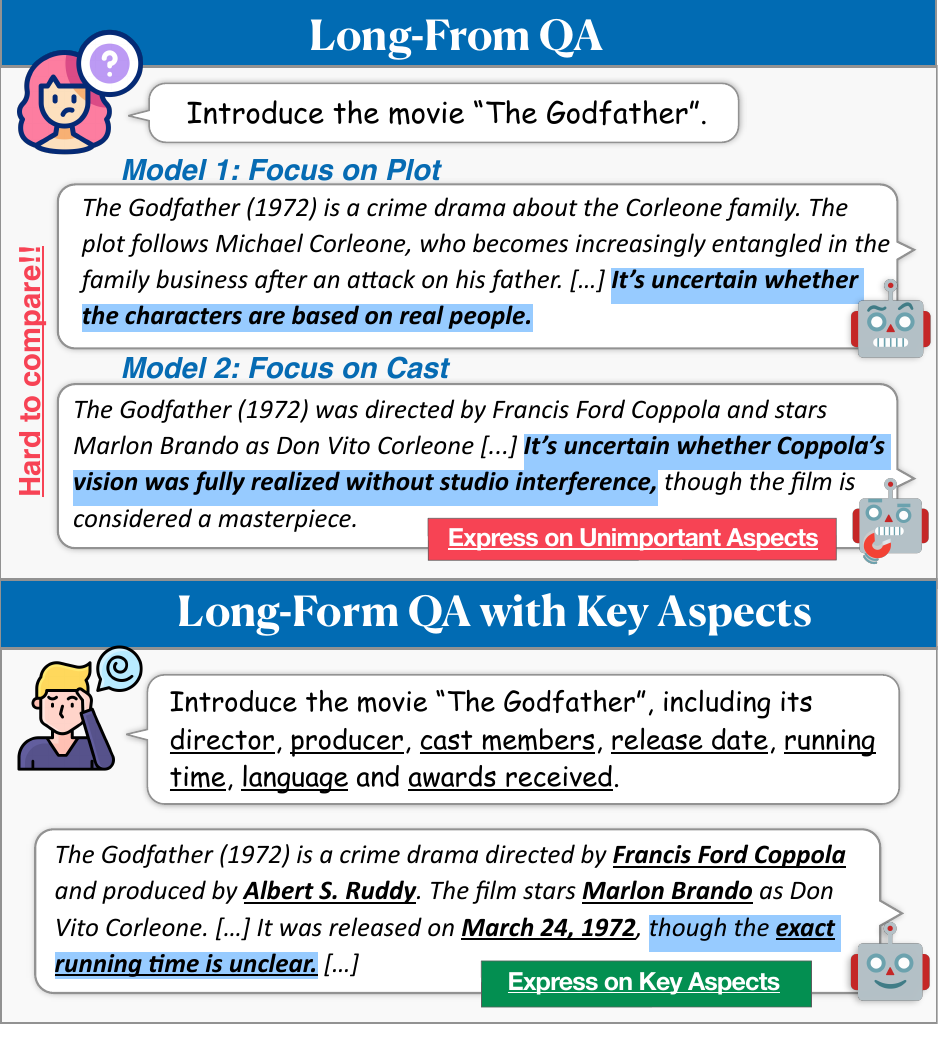}
    \caption{Evaluating uncertainty in long-form generation is challenging: different models may express uncertainty across varying aspects, often focusing on less important ones. Restricting the \textbf{key aspects} in long-form generation helps \textbf{ensure more consistent evaluation}.}
    \label{fig:problem}
\end{figure}

Large Language Models (LLMs) exhibit strong text generation abilities across diverse tasks and domains. However, they often hallucinate by generating incorrect or fabricated information \citep{zhang2023siren, huang2023survey}, especially when lacking sufficient knowledge \citep{gekhman-etal-2024-fine, li2023haluevallargescalehallucinationevaluation}. Enabling models to either refuse to answer or explicitly express uncertainty has emerged as a promising direction to reduce hallucinations and enhance trustworthiness \citep{zhang-etal-2024-luq, zhang2024atomic, logu}.

Current research on uncertainty expression in LLMs focuses primarily on short-form QA, where responses typically contain fewer than ten words \citep{kuhn2022semantic, lin2023generating, fadeeva-etal-2023-lm, wang2024sampleidentifygeneralframework}. However, real-world applications often require much longer outputs that may contain a mixture of correct and incorrect statements \citep{zhang-etal-2024-luq, huang-etal-2024-calibrating}. The challenge of \textit{estimating uncertainty in long-form generation} remains under-explored. 

Unlike previous post-hoc methods for long-form uncertainty estimation \citep{fadeeva-etal-2023-lm, zhang-etal-2024-luq, huang-etal-2024-calibrating, jiang2024graphbased}, which provide numerical estimates of output uncertainty, we explore the use of \textbf{linguistic uncertainty expressions} (\eg ``it is unclear whether'' or ``I am not sure''). These expressions are generated \textit{along with the output responses in a single decoding pass} to convey uncertainty or lack of knowledge \citep{zhou-etal-2023-navigating, kim2024m}. We argue that such explicit and human-interpretable expressions not only align more closely with daily communication but also offer efficiency advantages, as they are produced on-the-fly with minimal additional computational cost.

Regarding linguistic uncertainty expression in long-form generation, \citet{logu} propose a two-stage training approach to address uncertainty suppression and alignment issues. \citet{band2024linguistic} introduce linguistic calibration, enabling models to express uncertainty at different levels (\eg I am 70\% sure). However, due to the open-ended nature of long-form QA, different models may focus on different aspects and express uncertainty from different angles, making direct comparison challenging (upper; Figure \ref{fig:problem}). As a result, prior work does not answer a key research question: \textit{How can we fairly evaluate different models' ability to accurately express uncertainty in long-form generation?}

In this work, we introduce \uncle (\textbf{Unc}ertainty in \textbf{L}ong-form \textbf{E}xpressions), the first benchmark designed to comprehensively evaluate a model's ability to accurately express uncertainty in both long-form and short-form generation (Contribution \textbf{\#1}). \textbf{Our dataset directly bridges short- and long-form QA with paired questions and gold answers.} Each question contains one topic entity with multiple key aspects that models are expected to cover in their responses (Figure \ref{fig:problem} bottom; more examples in Table \ref{tab:uncle}). Each aspect is associated with a short-form question and a ground truth answer. The dataset spans five domains (biographies, companies, movies, astronomical objects, and diseases), containing over 1{,}000 entities. 
We also propose a suite of novel metrics to provide a comprehensive evaluation of uncertainty expression (Section \ref{sec:benckmark}). 

Using \uncle as a unified testbed, we evaluate ten popular LLMs to assess their ability to accurately express uncertainty in long-form generation. We reveal three key findings (Contribution \textbf{\#2}): 
(1) Although models can generally provide correct answers for known facts, current models show limited ability to accurately express uncertainty for unknown facts. (2) Closed-source models tend to use uncertainty expressions more frequently, while open-source models express uncertainty more accurately. (3) Models are more likely to use uncertainty expressions in short-form QA than in long-form QA (Section~\ref{sec:original}). 

Given that \uncle provides a direct comparison between short- and long-form uncertainty expressions, we investigate strategies to enhance model performance in both formats (Contribution \textbf{\#3}; Section~\ref{sec:training}). We consider both prompt-based and training-based approaches. We experiment with various training settings: exclusively short-form QA, exclusively long-form QA, and a mixture of both. Our results demonstrate that both prompt-based and training-based approaches improve over the base model, with training-based methods generally achieving greater gains. Meanwhile, training on long-form tasks benefits short-form tasks, but not vice versa. Furthermore, we analyze the alignment between short- and long-form uncertainty expressions and reveal a significant alignment gap (Section~\ref{sec:discussion}). We encourage future research to develop methods with \uncle that perform robustly across both QA formats.
\section{Related Work}

\begin{table*}[ht!]
\centering
\scriptsize
\begin{tabularx}{\textwidth}{>{\centering\arraybackslash}m{1.6cm}>{\centering\arraybackslash}m{1.9cm}>{\centering\arraybackslash}m{4.6cm}>{\centering\arraybackslash}m{4.6cm}>{\centering\arraybackslash}m{1.2cm}}
\toprule
\rowcolor[gray]{0.95}\textbf{Domains} & \textbf{Entities} & \textbf{Long-form QA Example} & \textbf{Short-form QA Example} & \textbf{\# Entities} \\
\midrule
Bios & Jackie Chan, Eminem, Steve Jobs... & In a paragraph, introduce the person Jackie Chan, including birthdate, place of birth, citizenship, language spoken, ... & What is Jackie Chan's birthdate? Where was Jackie Chan born? What is Jackie Chan's citizenship? ... & 319  \\
\midrule
Companies & Amazon, JP Morgan, Mars Incorporated... & In a paragraph, introduce the company Amazon, including date of establishment, founders, location of formation, CEO, ... & When was Amazon established? Who are the founders of Amazon? Where was Amazon formed? ... & 264 \\
\midrule
Movies & The Matrix, Inception, Fight Club... & In a paragraph, introduce the movie The Matrix, including genre, director, publication date, duration... & What is the genre of The Matrix? Who directed The Matrix? When was The Matrix first released? ... & 236 \\
\midrule
Astronomical Objects & Pluto, Uranus, Saturn... & In a paragraph, introduce the astronomical object Pluto, including mass, radius, orbital period, density... & What is Pluto's mass? What is Pluto's radius? What is Pluto's orbital period? What is Pluto's density? & 171 \\
\midrule
Diseases & HIV/AIDS, Tuberculosis, PTSD... & In a paragraph, introduce the disease HIV/AIDS, including time of discovery, symptoms, medical examination, possible treatments... & When was HIV/AIDS discovered? What are the symptoms of HIV/AIDS? How is HIV/AIDS diagnosed? ... & 76 \\
\midrule
\multicolumn{4}{r}{\textbf{In Total:}} & \textbf{1066} \\
\bottomrule
\end{tabularx}
\caption{Overview of the \uncle benchmark. Each entity is associated with multiple key aspects, which are formulated as both long-form and short-form questions. For the same entity, there could be many different questions covering different aspects. 
}
\label{tab:uncle}
\end{table*}

\rparagraph{Evaluating Long-form Factuality and Uncertainty} The evaluation of factuality in long-form generation has been extensively studied \citep{min-etal-2023-factscore, wei2024longfact, zhao2024wildhallu, song2024veriscore, chiang-lee-2024-merging}, typically by decomposing the text into atomic claims and verifying each claim using external knowledge sources. Existing LLMs have demonstrated strong performance in generating and verifying atomic claims, achieving low error rates compared to human annotation \citep{min-etal-2023-factscore, zhang-etal-2024-luq}. However, none of these studies specifically examine whether model-generated responses \textit{contain uncertainty expressions or whether those expressions are accurate}. 
On the other hand, existing studies on estimating uncertainty in long-form generation primarily focus on post-hoc methods \citep{zhang-etal-2024-luq, zhang2024atomic, huang-etal-2024-calibrating, jiang2024graphbased}, where a confidence score is assigned to each response, and traditional metrics like Spearman correlation or AUROC are used for a response level evaluation. Limited work has been done to assess how accurately models express uncertainty in long-form generation for each claim.

\rparagraph{Training LLMs to Express Uncertainty}
Most existing approaches for training language models to express uncertainty focus on short-form responses, where uncertainty is expressed about a single aspect. Several methods~\citep{xu2024sayself, zhang-etal-2024-r, han2024enhancingconfidenceexpressionlarge, lin2022teaching, madaan2023self} employ a two-stage strategy: first, the model answers the question, and then it is prompted again to provide a confidence label for the answer.  
Another line of work~\citep{cheng2024aiassistantsknowdont, chen2024teachinglargelanguagemodels, li2024know, wang2025sconuselectiveconformaluncertainty} encourages models to explicitly state ``I don't know'' when faced with unknown information, instead of generating incorrect answers with low-confidence.  

Teaching models to express uncertainty in long-form responses remains challenging due to the complexity of handling mixed uncertainties across multiple aspects in open-ended questions. Existing short-form QA methods do not transfer directly to the long-form setting. Recent work has explored this challenge. LoGU~\citep{logu} identifies two challenges in long-form uncertainty expression: uncertainty suppression and uncertainty misalignment. The authors then propose a two-step training framework: first, supervised fine-tuning to mitigate uncertainty suppression in long-form responses, followed by preference learning to address uncertainty misalignment. Linguistic Calibration~\citep{band2024linguistic} explores the feasibility of assigning a numerical confidence score to statements during generation. However, both approaches overlook a key issue: \textit{different models may produce different answers}, and each answer may express uncertainty from different angles. This variability hinders direct comparison of models' uncertainty expression.
\section{\uncle Construction} \label{sec:benckmark}
\subsection{Motivation}

Evaluating uncertainty expression in long-form generation is challenging due to the open-ended nature of existing long-form QA datasets. Most existing datasets \citep{min-etal-2023-factscore, wei2024longfact, zhao2024wildhallu} focus on questions regarding a single specific topic (\eg a person or an event) and prompt models to generate information broadly related to a topic entity (\eg ``Tell me a biography of \texttt{[PERSON]}''). 
Due to this openness, \textbf{any relevant details about the topic are generally accepted, making uncertainty evaluation difficult.} This open-endedness raises two key issues: 
\begin{inparaenum}[\it 1)]
\item models may express uncertainty in \textit{different} aspects, complicating cross-model comparisons, and 
\item models often express uncertainty for \textit{unimportant} details. 
\end{inparaenum} 
As shown in Figure \ref{fig:problem}, given the question ``Introduce the movie \textit{The Godfather},'' different models may emphasize various aspects, such as the plot, cast, or the film’s impact and awards, complicating fair comparisons across models. These challenges motivate the construction of a dataset \textbf{requiring long-form generation while maintaining relatively fixed answer aspects}. 
Specifically, we propose that models must cover several key aspects within their responses, maintaining the long-form nature of answers while improving coherence and comparability. Formally, for a question $q$ about an entity $e$, we define a set of key aspects $\mathcal{A}$ that must be included in the final answer. In the earlier example, the new question would be: ``Introduce the movie \textit{The Godfather}, including its director, producer, cast members, release date, running time, language, and awards received.''

\begin{table}[t] 
\setlength{\tabcolsep}{4pt} 
\renewcommand{\arraystretch}{0.95}
\footnotesize  
\centering
\resizebox{\columnwidth}{!}{
\begin{tabular}{lccc}
\toprule
\textbf{Dataset} & \textbf{Short-form} & \textbf{Long-form} & \textbf{Gold Ans.} \\
\midrule
TriviaQA~\citep{joshi2017triviaqalargescaledistantly} & \checkmark &  & \checkmark \\
Natural Questions~\citep{kwiatkowski-etal-2019-natural} & \checkmark &  & \checkmark \\
SimpleQA~\citep{wei2024measuringshortformfactualitylarge} & \checkmark &  & \checkmark \\
FactScore~\citep{min2023factscorefinegrainedatomicevaluation} &  & \checkmark &  \\
LongFact~\citep{wei2024longformfactualitylargelanguage} &  & \checkmark &  \\
WildHallu~\citep{zhao2024wildhallucinationsevaluatinglongformfactuality} &  & \checkmark &  \\
\midrule
\uncle\ (\textbf{Ours}) & \checkmark & \checkmark & \checkmark \\
\bottomrule
\end{tabular}}
\caption{Comparison between \uncle and other popular datasets in uncertainty estimation.}
\label{tab:dataset_comparison}
\vspace{-8pt}
\end{table}

\subsection{Data Collection}
To collect the questions in our dataset, we need the entities $\mathcal{E}$, key aspects $\mathcal{A}$, and a knowledge base $\mathcal{C}$. We adopt Wikidata as the source for all $\mathcal{E}$, $\mathcal{A}$, and $\mathcal{C}$. Wikidata consists of knowledge triplets in the form of (Subject, Predicate, Object). We use the \textit{predicates} in Wikipedia to construct our \textit{aspects}.
Our dataset spans five domains: biographies, companies, films, astronomical objects, and diseases. Each domain uses a tailored set of aspects. We outline the construction procedure below.

\rparagraph{Step 1: Sampling Entities $\mathcal{E}$} For each domain, we select entities from Wikidata spanning different categories and frequencies. Following \citet{liu2024multi}, we use the number of properties associated with an entity as a proxy for its frequency, which serves as an indicator of the amount of information available online for that entity. This approach ensures a diverse set of entities with varying degrees of informational richness.

\rparagraph{Step 2: Sampling Key Aspects $\mathcal{A}$} We identify key aspects $\mathcal{A}$ for each domain by selecting the most important and relevant properties for answering questions. For instance, birth date and birthplace are essential for biographies, while founders and founding dates are crucial for companies. 

We retrieve and count the frequencies of all properties associated with \( \mathcal{E} \), and retain the most frequent ones. For \( \mathcal{E} \), we first retrieve all associated properties from Wikidata. For each property $P$, we count how many of the entities \( \mathcal{E} \) possess it, retaining only the most frequent properties as key aspects for each entity. This process ensures that the key aspects are representative and stable by filtering out rare properties, which might otherwise introduce noise or bias into the evaluation. Five distinct groups of key aspects are selected for our five domains, followed by human verification. 

\rparagraph{Step 3: Generating Questions} For each entity, we generate two types of questions: 
\begin{inparaenum}[\it 1)]
\item \textbf{Long-form:} These require comprehensive answers covering multiple key aspects in a coherent paragraph.
\item \textbf{Short-form:} Concise, fact-based questions targeting specific aspects. GPT-4o is prompted to generate questions, with ground-truth answers provided for each short-form question. We maintain a dataset of approximately 1k entities for affordability and usability. Each entity can yield multiple long-form questions by combining aspects.
\end{inparaenum}

Table~\ref{tab:uncle} reports dataset statistics and examples.
Table~\ref{tab:dataset_comparison} compares \uncle with prior work.
As shown in the table, \uncle is the only dataset that pairs short- and long-form questions with gold answers.
Details of the human annotation for quality verification are in Appendix~\ref{app:annotation}.

\section{Task Definition and Evaluation}

\begin{figure*}[t!]
    \centering
    \includegraphics[width=0.92\textwidth]{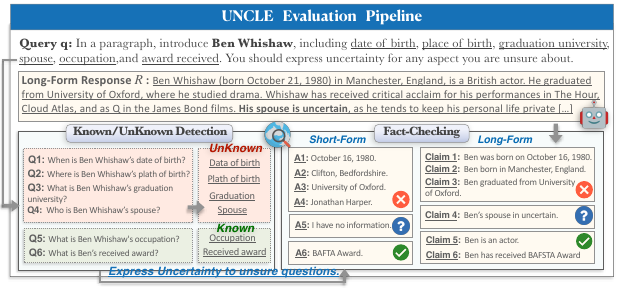}
    \caption{
    Evaluation Pipeline for \uncle. The framework consists of three steps: detecting known/unknown key aspects, generating long- and short-form answers, and fact-checking. \includegraphics[width=1.05em]{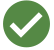} represents a correct answer, \includegraphics[width=1.05em]{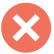} represents an incorrect answer, and \includegraphics[width=1.05em]{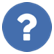} represents uncertainty expression.
}
    \vspace{-2mm}
    \label{fig:main}
\end{figure*}

\begin{table}[t!]
\setlength\tabcolsep{4pt}
\renewcommand{\arraystretch}{1.3} 
\centering
\footnotesize
\begin{tabular}{l | >{\centering\arraybackslash}p{1.25cm} >{\centering\arraybackslash}p{1.25cm} >{\centering\arraybackslash}p{1.25cm} |c}

\toprule
& \textbf{Correct} & \textbf{Incorrect} & \textbf{Uncertain} &   \\
\hline
\textbf{Known}   &$\mathcal{A}_{{\text{kn}}}^{\text{cor}}$      
\cellcolor{darkgreen!50}&$\mathcal{A}_{\text{kn}}^{\text{incor}}$
\cellcolor[gray]{0.95}&$\mathcal{A}_{\text{kn}}^{\text{unc}}$      \cellcolor[gray]{0.95}
&$\mathcal{A}_{\text{kn}}$   \\
\textbf{Unknown}   &\cellcolor[gray]{0.95}$\mathcal{A}_{{\text{unk}}}^{\text{cor}}$ 
\cellcolor[gray]{0.95}&$\mathcal{A}_{\text{unk}}^{\text{incor}}$ 
\cellcolor[gray]{0.95}&$\mathcal{A}_{\text{unk}}^{\text{unc}}$   
\cellcolor{darkgreen!50}
&$\mathcal{A}_{\text{unk}}$   \\
\hline
&$ \mathcal{A}_{\text{cor}}$ 
&$ \mathcal{A}_{\text{incor}}$ 
&$\mathcal{A}_{\text{unc}}$ 
& \\ 
\bottomrule
\end{tabular}
\caption{Uncertainty confusion matrix. \texttt{Correct}, \texttt{Incorrect}, and \texttt{Uncertain} are based on the model's response, while \texttt{Known} and \texttt{Unknown} refer to the results of knowledge probing. Ideally, the model should correctly represent known facts and express uncertainty when faced with unknown facts, as highlighted in \colorbox{darkgreen!50}{green}.}
\label{tab:metric}
\end{table}

We define a long- and short-form generation task with restricted key aspects as follows. For an entity \( e \), its corresponding key aspects are denoted as \( \mathcal{A} =\cup_{i}{\mathcal{A}_i}\). 
For long-form QA, we construct a query \( q(e \mid \mathcal{A}) \), specifying the key aspects to cover (\eg "Introduce \texttt{[ENTITY]} to me, including \texttt{[A1]}, \texttt{[A2]}, \texttt{[A3]}, \dots"). 
We prompt language model \( \mathcal{M} \) with \( q(e \mid \mathcal{A}) \). The response is denoted as \( R \sim \mathcal{M}(R \mid q(e \mid \mathcal{A})) \). 
For short-form QA, we prompt $\mathcal{M}$ with individual questions for each aspect \( q(e \mid \mathcal{A}_i)\). The short-form response is denoted as \( R_i \sim \mathcal{M}(R_i \mid \mathcal{A}_i) \).

\rparagraph{Known/Unknown Detection} For a specific LLM \( \mathcal{M} \), we categorize the aspects \( \mathcal{A}_i \) into two groups based on the model's knowledge: known aspects \( \mathcal{A}_{\text{kn}} \) and unknown aspects \( \mathcal{A}_{\text{unk}} \). For knowledge probing, we follow previous work \citep{gekhman-etal-2024-fine, yang2023alignment} to query the model multiple times; if \( \mathcal{M} \) consistently fails to provide correct answers, the corresponding knowledge is regarded as unknown.

\rparagraph{Response Categorization} The response \( R \) is expected to include information about the key aspects of \( \mathcal{A} \). These aspects are divided into three subsets based on correctness: \( \mathcal{A}_{\text{cor}} \) for correctly answered aspects, \( \mathcal{A}_{\text{incor}} \) for incorrectly answered aspects, and \( \mathcal{A}_{\text{unc}} \) for aspects where the model expresses uncertainty. We follow the same categorization for short-form responses \( R_i \).

We then construct the uncertainty confusion matrix shown in Table~\ref{tab:metric}. In our setting, existing metrics such as AUROC and ECE are not applicable, as our linguistic uncertainty level is binary rather than continuous. Therefore, we propose a suite of new evaluation metrics to comprehensively assess the model's ability to express uncertainty:

\begin{metric}[Factual Accuracy]
Let \( \mathcal{A}_{\rm{cor}} \) denote the set of correct aspects, and \( \mathcal{A}_{\rm{incor}} \) denote the set of incorrect aspects in the response. 
The \textbf{Factual Accuracy (FA)} is then defined as
\begin{equation*}
    FA = \frac{|\mathcal{A}_{\rm{cor}}|}{|\mathcal{A}_{\rm{cor}}| + |\mathcal{A}_{\rm{incor}}|}.
\end{equation*}
\end{metric}
FA measures the proportion of aspects that are stated correctly among all aspects that are stated certainly.

\begin{metric}[Uncertain Accuracy]
Let \( \mathcal{A}_{\text{\rm{unc}}} \) denote the set of aspects answered with uncertainty, and \( \mathcal{A}_{\text{\rm{unk}}}^{\rm{unc}} \) denote the set of unknown aspects within \( \mathcal{A}_{\text{\rm{unc}}} \). 
The \textbf{Uncertain Accuracy (UA)} is then defined as
\begin{equation*}
    \text{UA} = \frac{|\mathcal{A}_{\text{\rm{unk}}}^{\rm{unc}}|}{|\mathcal{A}_{\text{\rm{unc}}}|}.
\end{equation*}
\end{metric}
UA calculates how often the model accurately expresses uncertainty, \ie among the aspects the model expresses with uncertainty, the fraction that are truly unknown.

\begin{metric}[Known to Correct Rate]
Let \( \mathcal{A}_{\rm{kn}} \) denote the set of all known aspects, and \( \mathcal{A}_{\rm{kn}}^{\rm{cor}} \) denote the set of known aspects answered correctly. The \textbf{Known to Correct Rate (KCR)} is then defined as
\begin{equation*}
    \text{KCR} = \frac{|\mathcal{A}_{\rm{kn}}^{\rm{cor}}|}{|\mathcal{A}_{\rm{kn}}|}.
\end{equation*}
\end{metric}
KCR measures the proportion of aspects known to the model that are correctly expressed in the generated response. 

\begin{metric}[Unknown to Uncertain Rate] 
Let \( \mathcal{A}_{\rm{unk}} \) denote the set of all unknown aspects, and \( \mathcal{A}_{\rm{unk}}^{\rm{unc}} \) denote the set of unknown aspects expressed as uncertainty. The \textbf{Unknown to Uncertain Rate (UUR)} is then defined as
\begin{equation*}
    \text{UUR} = \frac{|\mathcal{A}_{\rm{unk}}^{\rm{unc}}|}{|\mathcal{A}_{\rm{unk}}|}.
\end{equation*}
\end{metric}
UUR measures the proportion of aspects the model does not know that are expressed with uncertainty rather than incorrectly stated as facts.

\begin{metric}[Expression Accuracy]
With previously defined notations, \textbf{Expression Accuracy (EA)} is then defined as
\begin{equation*}
    \text{EA} = \frac{|\mathcal{A}_{\rm{kn}}^{\rm{cor}}| + |\mathcal{A}_{\rm{unk}}^{\rm{unc}}|}{|\mathcal{A}_{\rm{kn}}| + |\mathcal{A}_{\rm{unk}}|}.
\end{equation*}
\end{metric}
EA is the micro-average of KCR and UUR, quantifying the proportion of aspects that are correctly expressed, \ie the model maintains correct expressions for aspects it knows and expresses aspects it does not know as uncertainty.

\rparagraph{Metric Summary}
Taken together, these five metrics form a complementary evaluation suite, each capturing a distinct facet of uncertainty expression that existing metrics do not adequately capture. Factual Accuracy (FA) evaluates the correctness of \emph{confident statements}; Uncertain Accuracy (UA) checks whether uncertainty is expressed appropriately when the underlying aspect is unknown; Known to Correct Rate (KCR) measures the fraction of \emph{known} aspects stated correctly; Unknown to Uncertain Rate (UUR) measures the fraction of \emph{unknown} aspects explicitly marked as uncertain; and Expression Accuracy (EA) provides an overall measure of appropriate expression.

\rparagraph{Evaluation Pipeline}
\label{sec:eval}
Figure \ref{fig:main} illustrates the overview of our evaluation pipeline. 
\textbf{Step 1: Known/Unknown Detection.} To assess whether the model \textit{knows} a key aspect, we prompt it five times with the corresponding short-form question at a temperature of 1 \citep{yang2023alignment, gekhman-etal-2024-fine}. If none of the five responses are correct, we classify the aspect as unknown; otherwise, it is considered known. 
\textbf{Step 2: Question Answering.} We then prompt the model to answer both short- and long-form questions with temperature 0. In the prompt, we explicitly ask the model to express uncertainty.
\textbf{Step 3: Fact-checking.} We first collect all answers where the models express uncertainty, using GPT-4o \citep{gpt4o}. For the remaining certain answers, we use GPT-4o to compare them against a gold reference for each key aspect. Each aspect is then classified as correct, incorrect, or uncertain.
\textbf{Step 4: Calculating Metrics.} We then draw the confusion matrix in Table \ref{tab:metric} and calculate our five metrics.
We also perform a human evaluation (see Appendix~\ref{app:annotation}) to verify the reliability of our automated assessment pipeline. All prompts are listed in Appendix~\ref{sec:logu_prompt}.
\definecolor{Gray}{gray}{0.92}
\begin{table*}[t!]
\footnotesize
\centering
\setlength\tabcolsep{9pt}
\scalebox{0.9}[0.9]{ 
\begin{tabular}{lcccccccccc}
\toprule
\multirow{2}{*}{\textbf{Method}} & \multicolumn{5}{c}{\textbf{Long-Form}} & \multicolumn{5}{c}{\textbf{Short-Form}} \\
\cmidrule(lr){2-6} \cmidrule(lr){7-11}
& \textbf{$\mathbf{FA}$}$\uparrow$
& \textbf{$\mathbf{UA}$}$\uparrow$
& \textbf{$\mathbf{UUR}$}$\uparrow$ 
& \textbf{$\mathbf{KCR}$}$\uparrow$
& \textbf{$\mathbf{EA}$}$\uparrow$
& \textbf{$\mathbf{FA}$}$\uparrow$
& \textbf{$\mathbf{UA}$}$\uparrow$
& \textbf{$\mathbf{UUR}$}$\uparrow$ 
& \textbf{$\mathbf{KCR}$}$\uparrow$
& \textbf{$\mathbf{EA}$}$\uparrow$ \\
\midrule
\multicolumn{11}{c}{\textbf{\textit{Close-sourced Models}}} \\
\midrule
GPT-3.5 & \cellcolorval{73.7} & \cellcolorval{32.3} & \cellcolorval{2.08} & \cellcolorval{87.9} & \cellcolorval{66.8}  & \cellcolorval{76.7} &\cellcolorval{2.63} & \cellcolorval{0.54} & \cellcolorval{97.4} &\cellcolorval{74.7}  \\
GPT-4 & \cellcolorval{76.9} & \cellcolorval{13.8} & \cellcolorval{6.00} & \cellcolorval{87.2} & \cellcolorval{74.8} & \cellcolorval{84.2} &\cellcolorval{4.82} & \cellcolorval{3.11} & \cellcolorval{95.1} &\cellcolorval{81.1} \\
Claude-3.5-Sonnet & \cellcolorval{75.3} & \cellcolorval{8.25} & \cellcolorval{0.97} & \cellcolorval{96.1} & \cellcolorval{72.0} 
& \cellcolorval{86.7} &\cellcolorval{3.08} & \cellcolorval{2.76} & \cellcolorval{97.4}  &\cellcolorval{84.7} \\
DeepSeek-Chat & \cellcolorval{73.7} & \cellcolorval{29.2} & \cellcolorval{0.83} & \cellcolorval{94.5} & \cellcolorval{70.6} 
& \cellcolorval{78.6} &\cellcolorval{7.45} & \cellcolorval{2.90} & \cellcolorval{95.5}  &\cellcolorval{76.7} \\
\midrule
\multicolumn{11}{c}{\textbf{\textit{Open-sourced Models}}} \\
\midrule
Llama-3-8B & \cellcolorval{58.0} & \cellcolorval{41.2} & \cellcolorval{1.12} & \cellcolorval{81.6} & \cellcolorval{50.7}  
& \cellcolorval{63.3} &\cellcolorval{42.3} & \cellcolorval{2.42} & \cellcolorval{89.2} &\cellcolorval{55.8}  \\
Llama-3-70B & \cellcolorval{70.2} & \cellcolorval{40.0} & \cellcolorval{0.79} & \cellcolorval{85.4} & \cellcolorval{65.8}  
& \cellcolorval{75.2} &\cellcolorval{25.0} & \cellcolorval{1.86} & \cellcolorval{92.5}  &\cellcolorval{71.4} \\
Mixtral-7B & \cellcolorval{52.7} & \cellcolorval{46.7} & \cellcolorval{2.16} & \cellcolorval{78.9} & \cellcolorval{46.9} 
& \cellcolorval{58.8} &\cellcolorval{25.9} & \cellcolorval{3.47} & \cellcolorval{89.7} &\cellcolorval{53.8}  \\
Mixtral-8x7B & \cellcolorval{66.3} & \cellcolorval{37.2} & \cellcolorval{1.87} & \cellcolorval{83.4} & \cellcolorval{61.0}  
& \cellcolorval{72.9} &\cellcolorval{22.3} & \cellcolorval{5.41} & \cellcolorval{92.6} & \cellcolorval{68.8} \\
Qwen2-7B & \cellcolorval{48.7}  & \cellcolorval{57.8} & \cellcolorval{4.00} & \cellcolorval{79.9} & \cellcolorval{41.4}
& \cellcolorval{47.8} &\cellcolorval{31.2} & \cellcolorval{4.05} & \cellcolorval{85.5} &\cellcolorval{43.1}  \\
Qwen2-72B & \cellcolorval{63.2}  & \cellcolorval{44.3} & \cellcolorval{2.78} & \cellcolorval{84.7} & \cellcolorval{58.8}
& \cellcolorval{68.9} &\cellcolorval{22.8} & \cellcolorval{4.75} & \cellcolorval{92.2} &\cellcolorval{64.5}  \\
\bottomrule
\end{tabular}
}
\caption{Performance of Different Models on \uncle. All values are presented as percentages, with darker colors representing higher scores. Metrics include Factual Accuracy (FA), Uncertain Accuracy (UA), Known to Correct Rate (KCR), Unknown to Uncertain Rate (UUR), and Expression Accuracy (EA).}
\label{tab:RQ1}
\end{table*}

\section{LLMs' Performance on \uncle} \label{sec:original}
Leveraging \uncle, we first explore the following question: \textit{How well do current LLMs selectively express uncertainty in long-form generation?}

\subsection{Models and Prompts}
We conduct our experiments with both open- and closed-sourced models: GPT-3.5-turbo-1106~\citep{gpt3.5}, GPT-4-1106-preview~\citep{gpt4o}, Claude-3.5-Sonnet~\citep{anthropic2023claude}, Deepseek-Chat~\citep{deepseekai2025deepseekv3technicalreport}, Llama3 Instruct (8B and 70B) \citep{llama3modelcard}, Mistral Instruct (7B and 8x7B)~\citep{jiang2023mistral}, and Qwen2 Instruct (7B and 72B)~\citep{yang2024qwen2technicalreport}.
For both long-form and short-form generation, the model is directly prompted to express uncertainty with ``You should express uncertainty for any aspect you are unsure about.'' (see full prompt in Appendix~\ref{sec:logu_prompt}). 

\subsection{Results}

\rparagraph{Models exhibit consistently low UA and UUR in both long- and short-form QA} 
As shown in Table~\ref{tab:RQ1}, all models are with UUR below 10\%, indicating a limited ability to express unknown cases through uncertainty expressions. UA generally remains below 50\%, and open-source models perform generally better. A closer analysis reveals that open-source models tend to \textit{produce more uncertainty expressions}, resulting in a larger $\mathcal{A}_{unc}$. However, many of these expressions do not correspond to truly unknown cases, leading to lower UA. In contrast, closed-source models produce fewer uncertainty expressions but do so more accurately, resulting in higher UA. Overall, current models struggle to express uncertainty  accurately in both long- and short-form QA.

\rparagraph{Models achieve relatively high KCR and EA across QA formats} 
All models exceed 75\% KCR on long-form QA and 85\% on short-form QA, indicating strong performance on correctly stating known knowledge. Notably, the models with the highest KCR also achieve the highest EA. This is because EA rewards both correct answers to known questions (KCR) and appropriate handling of unknowns (UUR). Ideally, models should excel in both KCR and UUR, but current performance on UUR remains inadequate.

\rparagraph{Short-form QA yields higher FA, KCR, and EA, but lower UA compared to long-form QA} 
A closer examination reveals that models tend to express uncertainty more frequently in short-form QA, resulting in a larger $\mathcal{A}_{unc}$. However, many of these expressions do not correspond to truly unknown cases, which lowers UA. Meanwhile, the higher FA observed in short-form settings is likely due to the narrower scope of each question, which reduces noise and improves factual accuracy. Further analysis is presented in \S\ref{sec:discussion}.

\section{Teaching LLMs to Express Uncertainty} \label{sec:training}

\definecolor{Gray}{gray}{0.92}
\begin{table*}[t!]
\centering
\footnotesize
\setlength\tabcolsep{9.5pt}
\scalebox{0.88}[0.88]{ 
\begin{tabular}{llcccccccccc}
\toprule
& \multirow{2}{*}{\textbf{Method}} & \multicolumn{5}{c}{\textbf{Long-Form}} & \multicolumn{5}{c}{\textbf{Short-Form}} \\
\cmidrule(lr){3-7} \cmidrule(lr){8-12}
&
& \textbf{$\mathbf{FA}$}$\uparrow$
& \textbf{$\mathbf{UA}$}$\uparrow$ 
& \textbf{$\mathbf{UUR}$}$\uparrow$
& \textbf{$\mathbf{KCR}$}$\uparrow$
& \textbf{$\mathbf{EA}$}$\uparrow$
& \textbf{$\mathbf{FA}$}$\uparrow$
& \textbf{$\mathbf{UA}$}$\uparrow$ 
& \textbf{$\mathbf{UUR}$}$\uparrow$
& \textbf{$\mathbf{KCR}$}$\uparrow$
& \textbf{$\mathbf{EA}$}$\uparrow$ \\
\midrule
\multicolumn{12}{c}{\textit{\textbf{Llama3-8B-Instruct}}} \\
\midrule
& Unc-Zero & \cellcolorval{58.0} & \cellcolorval{41.2} & \cellcolorval{1.12} & \cellcolorval{81.6} &\cellcolorval{50.7} &\cellcolorval{63.3} &\cellcolorval{42.3} &\cellcolorval{2.42} &\cellcolorval{89.2} &\cellcolorval{55.8} \\
\midrule
\multirow{3}{*}{\rotatebox{90}{Prompt}}
& Unc-Few & \cellcolorval{58.1} & \cellcolorval{64.0} & \cellcolorval{12.5} & \cellcolorval{75.8} &\cellcolorval{51.5} &\cellcolorval{72.4} &\cellcolorval{41.3} &\cellcolorval{86.2} &\cellcolorval{23.4} &\cellcolorval{47.6} \\
& Pair-Few & \cellcolorval{58.6} & \cellcolorval{51.1} & \cellcolorval{11.4} & \cellcolorval{75.8} &\cellcolorval{51.0} &\cellcolorval{69.1} &\cellcolorval{39.8} &\cellcolorval{92.1}  &\cellcolorval{13.9} &\cellcolorval{44.0}\\
& Self-Refine & \cellcolorval{53.8} & \cellcolorval{37.2} & \cellcolorval{12.4} & \cellcolorval{73.2} &\cellcolorval{47.8} &\cellcolorval{56.2} &\cellcolorval{34.9} &\cellcolorval{84.5}  &\cellcolorval{21.5} &\cellcolorval{47.6}\\
\midrule
\multirow{3}{*}{\rotatebox{90}{Training}}
&Short-DPO & \cellcolorval{56.7} & \cellcolorval{48.4} & \cellcolorval{13.4} & \cellcolorval{73.7} &\cellcolorval{50.5} &\cellcolorval{69.2} &\cellcolorval{62.6} &\cellcolorval{38.6} &\cellcolorval{79.5} &\cellcolorval{63.7} \\
&Mix-DPO &\cellcolorval{58.5} &\cellcolorval{56.1}  &\cellcolorval{34.2}  &\cellcolorval{65.7}  &\cellcolorval{53.6} &\cellcolorval{69.3}  &\cellcolorval{62.0}  &\cellcolorval{38.1} &\cellcolorval{79.4} &\cellcolorval{63.5} \\
&Long-DPO & \cellcolorval{51.5} & \cellcolorval{59.6} & \cellcolorval{40.7} & \cellcolorval{57.6} &\cellcolorval{51.1} &\cellcolorval{79.3} &\cellcolorval{55.0} &\cellcolorval{71.3} &\cellcolorval{61.0} &\cellcolorval{65.0} \\
\midrule
\multicolumn{12}{c}{\textit{\textbf{Mistral-7B-Instruct}}} \\
\midrule
& Unc-Zero & \cellcolorval{52.7} & \cellcolorval{46.7} & \cellcolorval{2.16} & \cellcolorval{78.9} & \cellcolorval{46.9} & \cellcolorval{58.8} & \cellcolorval{25.9} & \cellcolorval{3.47} &\cellcolorval{89.7} &\cellcolorval{53.8} \\
\midrule
\multirow{3}{*}{\rotatebox{90}{Prompt}}

& Unc-Few & \cellcolorval{54.9} & \cellcolorval{50.6} & \cellcolorval{7.00} & \cellcolorval{79.8} &\cellcolorval{49.5} &\cellcolorval{71.6} &\cellcolorval{53.5} &\cellcolorval{58.1} &\cellcolorval{67.5} &\cellcolorval{63.6} \\
& Pair-Few & \cellcolorval{54.2} & \cellcolorval{46.8} & \cellcolorval{2.77} & \cellcolorval{79.7} &\cellcolorval{47.7} &\cellcolorval{60.0} &\cellcolorval{45.4} &\cellcolorval{16.0} &\cellcolorval{83.6} &\cellcolorval{55.6} \\
& Self-Refine & \cellcolorval{41.7} & \cellcolorval{43.1} & \cellcolorval{8.60} & \cellcolorval{76.3} &\cellcolorval{42.5} &\cellcolorval{45.1} &\cellcolorval{42.6} &\cellcolorval{7.40} &\cellcolorval{76.8} &\cellcolorval{45.8} \\
\midrule
\multirow{3}{*}{\rotatebox{90}{Training}}
&Short-DPO & \cellcolorval{53.4} & \cellcolorval{51.9} & \cellcolorval{10.8} & \cellcolorval{79.5} &\cellcolorval{47.0} &\cellcolorval{69.8} &\cellcolorval{56.7} &\cellcolorval{45.8} &\cellcolorval{77.0} &\cellcolorval{64.1} \\
&Mix-DPO &\cellcolorval{56.9}  &\cellcolorval{54.6}  &\cellcolorval{43.1}  &\cellcolorval{61.2}  &\cellcolorval{53.7} &\cellcolorval{64.1} &\cellcolorval{54.9} &\cellcolorval{28.1} &\cellcolorval{81.9}  &\cellcolorval{59.5} \\
&Long-DPO & \cellcolorval{53.3} & \cellcolorval{59.6} & \cellcolorval{37.8} & \cellcolorval{62.2} &\cellcolorval{52.0} &\cellcolorval{70.1} &\cellcolorval{51.6} &\cellcolorval{58.1} &\cellcolorval{62.7} &\cellcolorval{60.8}\\
\bottomrule
\end{tabular}
}
\caption{Performance of Different Prompting and Training Strategies  on \uncle. All values are presented as percentages, with darker colors for higher scores. Metrics include Factual Accuracy (FA), Uncertain Accuracy (UA), Known to Correct Rate (KCR), Unknown to Uncertain Rate (UUR), and Expression Accuracy (EA).}
\label{tab:RQ2}
\end{table*}

We further explore both prompt-based and training-based methods to teach LLMs to express uncertainty in long-form generation (prompts and more training details are in Appendix \ref{app:teaching}. Detailed Analysis in Appendix \ref{app:category_analysis} and \ref{app:case_study}).

\subsection{Experiment Settings}
\label{sec:training_methods}
\rparagraph{Prompt-based Methods} 
\begin{inparaenum}[\it 1)]
    \item \textbf{Unc-Zero}: The model is directly prompted to express uncertainty in its output whenever it is unsure about any claims. This setting is identical to that used in Section~\ref{sec:original}.
    \item \textbf{Unc-Few}: Based on Unc-Zero, we provide the model with an additional set of 10 hand-crafted QA examples, where uncertainty is explicitly expressed in the answers as in-context learning examples. 
    \item \textbf{Pair-Few}: Extending Unc-Few, we provide the model with both a response containing only certain expressions, $R_{\text{cert}}$, and another with uncertainty expressions, $R_{\text{unc}}$, for each query. Each example is formatted as \texttt{<Q, $R_{\text{cert}}$, $R_{\text{unc}}$>}. The aim of including both $R_{\text{cert}}$ and $R_{\text{unc}}$ is to teach  models when to express uncertainty through in-context learning. \item \textbf{Self-Refine} \citep{madaan2023self}: We apply a draft-and-refine setup. The model is asked to first generate an initial response and then refine the uncertain claims into explicit uncertainty expressions in a second pass.
\end{inparaenum}

\rparagraph{Training-based Methods} 
We employ three training settings to teach the model to express uncertainty. Our \uncle dataset is used only for evaluation. 
\begin{inparaenum}[\it 1)]
    \item \textbf{Short-DPO}: Following \citet{cheng2024aiassistantsknowdont}, we conduct a two-stage SFT + DPO training using only short-form QA pairs.
    \item \textbf{Long-DPO}: Following \citet{logu}, we apply a similar two-stage SFT + DPO training approach, but using long-form QA examples exclusively.
    \item \textbf{Mix-DPO}: We mix training samples from Short-DPO and Long-DPO in a 3:7 ratio and perform two-stage training. To ensure fairness, the training datasets are kept the same size. Training Details can be found in Appendix \ref{sec:training_details}.
\end{inparaenum}

\subsection{Results}
\rparagraph{Both prompt- and training-based methods improve performance over Unc-Zero} 
We observe a substantial increase in UA and UUR, indicating improved capability in expressing uncertainty accurately. For instance, with Llama3, the UUR increases from 1.12\% under Unc-Zero to 34.2\% with Mix-DPO and 40.7\% with Long-DPO. Training-based methods generally yield greater improvements than prompt-based methods. Appendix~\ref{app:case_study} presents a case study comparing prompt-based and training-based methods.

\rparagraph{Training-based methods can better balance UUR and KCR} In contrast, the prompt-based methods tend to express excessive uncertainty, leading to a high UUR. For example, with the Llama3 in the short-form task, Pair-Few shows a high UUR of 92.1\% but a low KCR of 13.9\%. On the other hand, training-based methods like Long-DPO achieve a high UUR of 71.3\% while maintaining a high KCR of 61.0\%. The more balanced UUR and KCR also result in generally better EA compared to prompt-based methods. 

\rparagraph{Training on long-form tasks benefits short-form tasks, but not vice versa} 
For example, Llama3’s Long-DPO, trained on long-form tasks, achieves high UUR (71.3\%) and KCR (61.0\%) on short-form tasks. In contrast, Llama3’s Short-DPO performs poorly on long-form tasks, with a UUR of only 13.4\%. The Mix-DPO method offers a more balanced performance across both task formats. We hypothesize that training on long-form tasks, which involve multi-aspect uncertainty, enhances the model's ability to handle uncertainty in easier short-form settings.
\section{Discussion} \label{sec:discussion}
\subsection{Alignment Between Uncertainty Expressions in Short- and Long-form QA}

\begin{figure*}[t!]
    \centering
    \includegraphics[width=0.62\textwidth]{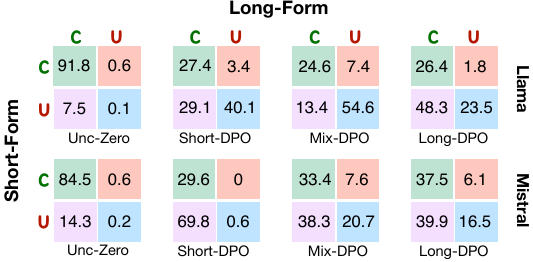}
    \caption{Distribution (in percentage) of key aspects expressed with certainty and uncertainty by Llama3 and Mistral across different training methods.
    \colorbox{CC}{C-C} indicates both short- and long-form express certainty, \colorbox{UU}{U-U} shows both express uncertainty, \colorbox{UC}{U-C} means uncertain in short-form but certain in long-form, and \colorbox{CU}{C-U} represents the reverse.}
    \vspace{-2mm}
    \label{fig:long-short}
\end{figure*}
Using paired short- and long-form questions in \uncle, we examine whether the same aspect is consistently expressed as certain or uncertain across different QA formats.
As shown in Figure~\ref{fig:long-short}, \colorbox{CC}{C-C} indicates the percentage of aspects expressed as certain in both short- and long-form QA, while \colorbox{UU}{U-U} represents those expressed as uncertain in both formats. Ideally, perfect alignment would result in all expressions falling into either \colorbox{CC}{C-C} or \colorbox{UU}{U-U}.

The key observations are as follows:
\begin{inparaenum}[\it 1)]
\item \textbf{In the original model (Unc-Zero), both short- and long-form aspects are mostly expressed with certainty.} \colorbox{CC}{C-C} accounts for 91.8\% in Llama3 and 84.5\% in Mistral, while both \colorbox{UU}{U-U} are below 1\%.
\item \textbf{Training increases the proportion of \colorbox{UU}{U-U} compared to Unc-Zero.} For Llama3, Short-DPO and Mix-DPO raise \colorbox{UU}{U-U} from 0.1\% to 40.1\% and 54.6\%, respectively.
\item \textbf{\colorbox{UC}{U-C} and \colorbox{CU}{C-U} remain substantial in training-based models.} This suggests ongoing inconsistency between short- and long-form uncertainty. Notably, \colorbox{UC}{U-C} often exceeds \colorbox{CU}{C-U}, indicating many aspects are certain in short-form QA but uncertain in long-form. Future work could improve alignment by reducing \colorbox{UC}{U-C} and \colorbox{CU}{C-U}.
\item \textbf{Trained only on short-form data, Llama3 and Mistral exhibit different ability in long-form QA.} For example, Mistral trained only on short-form data shows minimal long-form uncertainty (0\% + 0.6\%). In contrast, Llama3 retains long-form uncertainty even under the same condition (3.4\% + 40.1\%). This highlights the differing ability of models to generalize short-form uncertainty expression to long-form scenario.
\end{inparaenum}

\subsection{Influence of Mixture Ratio $\xi$}
\label{sec:analysis}

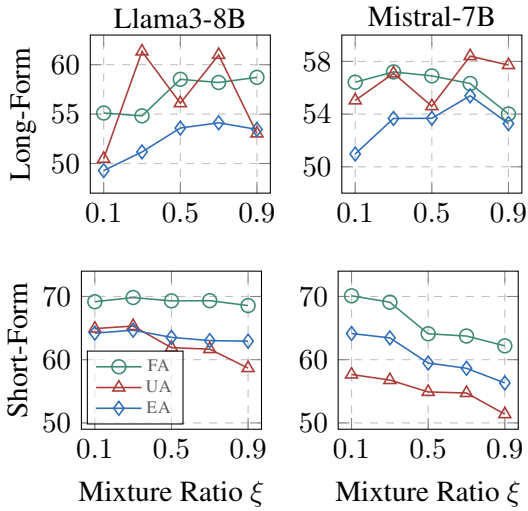
\begin{figure}[htbp]
    \vspace{-5pt}
    \centering
    \begin{minipage}[t]{0.225\textwidth}
        \centering
        \pgfplotsset{width=1.1\linewidth,height=1.02\linewidth,compat=1.5}
\begin{tikzpicture}
\begin{axis}[
    title={Llama3-8B},
    title style={yshift=-1.5ex},
    xmin=0.03, xmax=0.97,
    ymin=47, ymax=63,
    xtick={0.1, 0.5, 0.9},
    ytick={50, 55, 60},
    legend style={draw=none},
    ymajorgrids=true,
    xmajorgrids=true,
    grid style=dashed,
    ylabel={Long-Form},
    x label style={at={(axis description cs:0.5,-0.125)},anchor=north},
    y label style={at={(axis description cs:-0.125,0.5)},anchor=south, yshift=8pt},
]
\addplot[
    color=Green,
    mark=o,
    line width=0.6pt,
    mark size=2.6pt,
    ]
    coordinates {
    (0.1, 55.11)
    (0.3, 54.82)
    (0.5, 58.52)
    (0.7, 58.20)
    (0.9, 58.72)
    };

\addplot[
    color=Red,
    mark=triangle,
    line width=0.6pt,
    mark size=2.6pt,
    ]
    coordinates {
    (0.1, 50.48) 
    (0.3, 61.34)
    (0.5, 56.10)
    (0.7, 60.99) 
    (0.9, 53.04) 
    };
\addplot[
    color=Blue,
    mark=diamond,
    line width=0.6pt,
    mark size=2.6pt,
    ]
    coordinates {
    (0.1, 49.28) 
    (0.3, 51.17)
    (0.5, 53.59)
    (0.7, 54.12) 
    (0.9, 53.45) 
    };
\end{axis}
\end{tikzpicture}
        \label{fig:llama3-L}
    \end{minipage}
    \hspace{-1pt}
    \begin{minipage}[t]{0.225\textwidth}
        \centering
        \pgfplotsset{width=1.1\linewidth,height=1.02\linewidth,compat=1.5}
\begin{tikzpicture}
\begin{axis}[
    title={Mistral-7B},
    title style={yshift=-1.3ex},
    xmin=0.03, xmax=0.97,
    ymin=48, ymax=60,
    xtick={0.1, 0.5, 0.9},
    ytick={50, 54, 58},
    legend pos=south east,
    ymajorgrids=true,
    xmajorgrids=true,
    grid style=dashed,
    x label style={at={(axis description cs:0.5,-0.125)},anchor=north},
    y label style={at={(axis description cs:-0.125,0.5)},anchor=south},
    legend style={draw=none}
]
\addplot[
    color=Green,
    mark=o,
    line width=0.6pt,
    mark size=2.6pt,
    ]
    coordinates {
    (0.1, 56.42)
    (0.3, 57.20)
    (0.5, 56.90)
    (0.7, 56.32)
    (0.9, 54.01)
    };

\addplot[
    color=Red,
    mark=triangle,
    line width=0.6pt,
    mark size=2.6pt,
    ]
    coordinates {
    (0.1, 55.04) 
    (0.3, 57.09)
    (0.5, 54.60)
    (0.7, 58.40) 
    (0.9, 57.72) 
    };
\addplot[
    color=Blue,
    mark=diamond,
    line width=0.6pt,
    mark size=2.6pt,
    ]
    coordinates {
    (0.1, 50.97) 
    (0.3, 53.67)
    (0.5, 53.68)
    (0.7, 55.39) 
    (0.9, 53.28) 
    };
\end{axis}
\end{tikzpicture}
        \label{fig:llama3-S}
    \end{minipage}
    \vspace{-20pt}
    \begin{minipage}[t]{0.225\textwidth}
        \centering
        \pgfplotsset{width=1.1\linewidth,height=1.02\linewidth,compat=1.5}
\begin{tikzpicture}
\begin{axis}[
    xmin=0.03, xmax=0.97,
    ymin=49, ymax=74,
    xtick={0.1, 0.5, 0.9},
    ytick={50, 60, 70},
    legend pos=south west,
    ymajorgrids=true,
    xmajorgrids=true,
    grid style=dashed,
    xlabel={Mixture Ratio $\xi$},
    ylabel={Short-Form},
    x label style={at={(axis description cs:0.5,-0.125)},anchor=north, yshift=-8pt},
    y label style={at={(axis description cs:-0.125,0.5)},anchor=south, yshift=8pt},
    legend style={nodes={scale=0.6, transform shape},
    fill opacity=0.65,
    legend columns=1}
]
\addplot[
    color=Green,
    mark=o,
    line width=0.6pt,
    mark size=2.6pt,
    ]
    coordinates {
    (0.1, 69.17)
    (0.3, 69.83)
    (0.5, 69.30)
    (0.7, 69.33)
    (0.9, 68.56)
    };
    \addlegendentry{\textsc{FA}}

\addplot[
    color=Red,
    mark=triangle,
    line width=0.6pt,
    mark size=2.6pt,
    ]
    coordinates {
    (0.1, 64.92) 
    (0.3, 65.32)
    (0.5, 61.90)
    (0.7, 61.67) 
    (0.9, 58.67) 
    };
    \addlegendentry{\textsc{UA}}
\addplot[
    color=Blue,
    mark=diamond,
    line width=0.6pt,
    mark size=2.6pt,
    ]
    coordinates {
    (0.1, 64.20) 
    (0.3, 64.66)
    (0.5, 63.53)
    (0.7, 63.02) 
    (0.9, 62.93) 
    };
    \addlegendentry{\textsc{EA}}
\end{axis}
\end{tikzpicture}
        \label{fig:mistral-L}
    \end{minipage}
    \hspace{-1pt}
    \begin{minipage}[t]{0.225\textwidth}
        \centering
        \pgfplotsset{width=1.1\linewidth,height=1.02\linewidth,compat=1.5}
\begin{tikzpicture}
\begin{axis}[
    xmin=0.03, xmax=0.97,
    ymin=49, ymax=74,
    xtick={0.1, 0.5, 0.9},
    ytick={50, 60, 70},
    legend pos=south east,
    ymajorgrids=true,
    xmajorgrids=true,
    grid style=dashed,
    xlabel={Mixture Ratio $\xi$},
    x label style={at={(axis description cs:0.5,-0.125)},anchor=north, yshift=-8pt},
    y label style={at={(axis description cs:-0.125,0.5)},anchor=south},
    legend style={draw=none}
]
\addplot[
    color=Green,
    mark=o,
    line width=0.6pt,
    mark size=2.6pt,
    ]
    coordinates {
    (0.1, 70.10)
    (0.3, 69.09)
    (0.5, 64.10)
    (0.7, 63.74)
    (0.9, 62.19)
    };
\addplot[
    color=Red,
    mark=triangle,
    line width=0.6pt,
    mark size=2.6pt,
    ]
    coordinates {
    (0.1, 57.66) 
    (0.3, 56.78)
    (0.5, 54.90)
    (0.7, 54.73) 
    (0.9, 51.38) 
    };
\addplot[
    color=Blue,
    mark=diamond,
    line width=0.6pt,
    mark size=2.6pt,
    ]
    coordinates {
    (0.1, 64.14) 
    (0.3, 63.45)
    (0.5, 59.47)
    (0.7, 58.64) 
    (0.9, 56.32) 
    };
\end{axis}
\end{tikzpicture}
        \label{fig:mistral-S}
    \end{minipage}
    \vspace{-3pt}
    \caption{Performance of Llama3-8B and Mistral-7B across different mixture ratios.}
    \vspace{-9pt}
    \label{fig:mixture_ratio}
\end{figure}

We further analyze how the mixture ratio $\xi$ (proportion of long-form data) affects performance. Training data are constructed by mixing long-form and short-form data from ratio 0.1 to 0.9, while keeping the total amount of training data constant.

From Figure~\ref{fig:mixture_ratio}, we observe two key insights:
\begin{inparaenum}[\it 1)] 
    \item \textbf{Performance on long-form and short-form tasks is a trade-off}: Increasing the mixture ratio improves long-form performance but reduces short-form performance. The model trained with mixed data performs between Long-DPO and Short-DPO for both tasks.
    \item \textbf{While increasing the ratio $\xi$ consistently hurts the performance on short-form QA, it does not consistently improve long-form QA.} For example, in Figure~\ref{fig:mixture_ratio} (upper right), increasing $\xi$ from 0.1 to 0.9 leads to a 7.82\% drop in short-form EA for Mistral-7B. However, both long-form FA and EA first increase and then decline (lower right). Based on this trade-off, we select a ratio of 0.7 to achieve more balanced results.
\end{inparaenum}

\section{Conclusion}
We introduce \uncle, a benchmark for evaluating uncertainty in long- and short-form QA. Our experiments show that models struggle to express uncertainty in long-form generation. 
While our training method mitigates this issue, a misalignment persists in uncertainty expression between long- and short-form generation. Future work should focus on enhancing consistency across both forms.

\section*{Limitation}

\paragraph{Focus on verbalized linguistic expressions}
This work concentrates on verbalized linguistic expressions of uncertainty. Although \uncle\ can also be applied to post-hoc uncertainty estimation, we intentionally do not explore post-hoc methods in this paper, as our primary objective is to assess the model’s intrinsic ability to express what it does and does not know. We view post-hoc and verbalized approaches as \textit{complementary rather than competing}: post-hoc techniques may yield higher calibration in some settings, whereas verbalized expressions are simpler and more human-interpretable. Future work could extend \uncle\ to benchmark and integrate post-hoc estimators alongside verbalized cues.

\paragraph{Known and unknown detection} 
To assess the model's known and unknown knowledge, we employ a multiple sampling method. Increasing the number of sampling iterations could enhance the accuracy of the knowledge estimation.
Alternative approaches~\citep{gekhman-etal-2024-fine, yang2023alignment} may also be applicable. For example, one could apply a threshold of varying strictness in the sampling process to identify "Maybe Known" knowledge, or analyze the model’s hidden states to determine whether it possesses specific knowledge.

\paragraph{Robustness across generation types}
As discussed in \S\ref{sec:training} and shown in Table~\ref{tab:RQ2}, we have not yet identified an effective solution that performs well on both long- and short-form generation tasks. Future research could investigate this challenge more thoroughly.

\section*{Ethics Statement}
Our research adheres to strict ethical guidelines. We verified the licenses of all software and datasets used in this study to ensure full compliance with their terms. During the human annotation process, all annotators provided informed consent for their data to be included in the project. No privacy concerns have been identified. Additionally, we have conducted a thorough assessment of the project and do not anticipate any further risks.

\section*{Acknowledgement}
We thank Xiaochen Zhu, Chengzu Li, and Zhichao Yang for their proofreading and valuable comments on this paper. We also acknowledge the use of an icon from Flaticon\footnote{\url{https://www.flaticon.com}} and thank its creators for providing this visually appealing design.

\bibliography{custom,anthology}

\begin{thebibliography}{51}
\providecommand{\natexlab}[1]{#1}

\bibitem[{Anthropic(2023)}]{anthropic2023claude}
Anthropic. 2023.
\newblock Introducing claude 2.1.
\newblock Available from Anthropic: \url{https://www.anthropic.com/news/claude-2-1}.

\bibitem[{Band et~al.(2024)Band, Li, Ma, and Hashimoto}]{band2024linguistic}
Neil Band, Xuechen Li, Tengyu Ma, and Tatsunori Hashimoto. 2024.
\newblock \href {https://openreview.net/forum?id=rJVjQSQ8ye} {Linguistic calibration of long-form generations}.
\newblock In \emph{Forty-first International Conference on Machine Learning, {ICML} 2024, Vienna, Austria, July 21-27, 2024}. OpenReview.net.

\bibitem[{Bishop et~al.(2024)Bishop, Ananiadou, and Xie}]{bishop-etal-2024-longdocfactscore}
Jennifer~A. Bishop, Sophia Ananiadou, and Qianqian Xie. 2024.
\newblock \href {https://aclanthology.org/2024.lrec-main.941/} {{L}ong{D}oc{FACTS}core: Evaluating the factuality of long document abstractive summarisation}.
\newblock In \emph{Proceedings of the 2024 Joint International Conference on Computational Linguistics, Language Resources and Evaluation (LREC-COLING 2024)}, pages 10777--10789, Torino, Italia. ELRA and ICCL.

\bibitem[{Chen et~al.(2024)Chen, Liang, Wang, Liang, Xiao, Wei, Chen, Hao, Han, and Wang}]{chen2024teachinglargelanguagemodels}
Lida Chen, Zujie Liang, Xintao Wang, Jiaqing Liang, Yanghua Xiao, Feng Wei, Jinglei Chen, Zhenghong Hao, Bing Han, and Wei Wang. 2024.
\newblock \href {https://arxiv.org/abs/2406.10881} {Teaching large language models to express knowledge boundary from their own signals}.

\bibitem[{Cheng et~al.(2024)Cheng, Sun, Liu, Zhang, Yin, Li, Li, He, Chen, and Qiu}]{cheng2024aiassistantsknowdont}
Qinyuan Cheng, Tianxiang Sun, Xiangyang Liu, Wenwei Zhang, Zhangyue Yin, Shimin Li, Linyang Li, Zhengfu He, Kai Chen, and Xipeng Qiu. 2024.
\newblock \href {https://openreview.net/forum?id=girxGkdECL} {Can {AI} assistants know what they don't know?}
\newblock In \emph{Forty-first International Conference on Machine Learning, {ICML} 2024, Vienna, Austria, July 21-27, 2024}. OpenReview.net.

\bibitem[{Chiang and Lee(2024)}]{chiang-lee-2024-merging}
Cheng-Han Chiang and Hung-yi Lee. 2024.
\newblock \href {https://doi.org/10.18653/v1/2024.findings-acl.160} {Merging facts, crafting fallacies: Evaluating the contradictory nature of aggregated factual claims in long-form generations}.
\newblock In \emph{Findings of the Association for Computational Linguistics: ACL 2024}, pages 2734--2751, Bangkok, Thailand. Association for Computational Linguistics.

\bibitem[{De~Cao et~al.(2021)De~Cao, Aziz, and Titov}]{de2021editing}
Nicola De~Cao, Wilker Aziz, and Ivan Titov. 2021.
\newblock \href {https://doi.org/10.18653/v1/2021.emnlp-main.522} {Editing factual knowledge in language models}.
\newblock In \emph{Proceedings of the 2021 Conference on Empirical Methods in Natural Language Processing}, pages 6491--6506, Online and Punta Cana, Dominican Republic. Association for Computational Linguistics.

\bibitem[{DeepSeek-AI et~al.(2024)DeepSeek-AI, Liu, Feng, Xue, Wang, Wu, Lu, Zhao, Deng, Zhang, Ruan, Dai, Guo, Yang, Chen, Ji, Li, Lin, Dai, Luo, Hao, Chen, Li, Zhang, Bao, Xu, Wang, Zhang, Ding, Xin, Gao, Li, Qu, Cai, Liang, Guo, Ni, Li, Wang, Chen, Chen, Yuan, Qiu, Li, Song, Dong, Hu, Gao, Guan, Huang, Yu, Wang, Zhang, Xu, Xia, Zhao, Wang, Zhang, Li, Wang, Zhang, Zhang, Tang, Li, Tian, Huang, Wang, Zhang, Wang, Zhu, Chen, Du, Chen, Jin, Ge, Zhang, Pan, Wang, Xu, Zhang, Chen, Li, Lu, Zhou, Chen, Wu, Ye, Ye, Ma, Wang, Zhou, Yu, Zhou, Pan, Wang, Yun, Pei, Sun, Xiao, Zeng, Zhao, An, Liu, Liang, Gao, Yu, Zhang, Li, Jin, Wang, Bi, Liu, Wang, Shen, Chen, Zhang, Chen, Nie, Sun, Wang, Cheng, Liu, Xie, Liu, Yu, Song, Shan, Zhou, Yang, Li, Su, Lin, Li, Wang, Wei, Zhu, Zhang, Xu, Xu, Huang, Li, Zhao, Sun, Li, Wang, Yu, Zheng, Zhang, Shi, Xiong, He, Tang, Piao, Wang, Tan, Ma, Liu, Guo, Wu, Ou, Zhu, Wang, Gong, Zou, He, Zha, Xiong, Ma, Yan, Luo, You, Liu, Zhou, Wu, Ren, Ren, Sha, Fu, Xu, Huang, Zhang, Xie, Zhang, Hao,
  Gou, Ma, Yan, Shao, Xu, Wu, Zhang, Li, Gu, Zhu, Liu, Li, Xie, Song, Gao, and Pan}]{deepseekai2025deepseekv3technicalreport}
DeepSeek-AI, Aixin Liu, Bei Feng, Bing Xue, Bingxuan Wang, Bochao Wu, Chengda Lu, Chenggang Zhao, Chengqi Deng, Chenyu Zhang, Chong Ruan, Damai Dai, Daya Guo, Dejian Yang, Deli Chen, Dongjie Ji, Erhang Li, Fangyun Lin, Fucong Dai, Fuli Luo, Guangbo Hao, Guanting Chen, Guowei Li, H.~Zhang, Han Bao, Hanwei Xu, Haocheng Wang, Haowei Zhang, Honghui Ding, Huajian Xin, Huazuo Gao, Hui Li, Hui Qu, J.~L. Cai, Jian Liang, Jianzhong Guo, Jiaqi Ni, Jiashi Li, Jiawei Wang, Jin Chen, Jingchang Chen, Jingyang Yuan, Junjie Qiu, Junlong Li, Junxiao Song, Kai Dong, Kai Hu, Kaige Gao, Kang Guan, Kexin Huang, Kuai Yu, Lean Wang, Lecong Zhang, Lei Xu, Leyi Xia, Liang Zhao, Litong Wang, Liyue Zhang, Meng Li, Miaojun Wang, Mingchuan Zhang, Minghua Zhang, Minghui Tang, Mingming Li, Ning Tian, Panpan Huang, Peiyi Wang, Peng Zhang, Qiancheng Wang, Qihao Zhu, Qinyu Chen, Qiushi Du, R.~J. Chen, R.~L. Jin, Ruiqi Ge, Ruisong Zhang, Ruizhe Pan, Runji Wang, Runxin Xu, Ruoyu Zhang, Ruyi Chen, S.~S. Li, Shanghao Lu, Shangyan Zhou, Shanhuang
  Chen, Shaoqing Wu, Shengfeng Ye, Shengfeng Ye, Shirong Ma, Shiyu Wang, Shuang Zhou, Shuiping Yu, Shunfeng Zhou, Shuting Pan, T.~Wang, Tao Yun, Tian Pei, Tianyu Sun, W.~L. Xiao, Wangding Zeng, Wanjia Zhao, Wei An, Wen Liu, Wenfeng Liang, Wenjun Gao, Wenqin Yu, Wentao Zhang, X.~Q. Li, Xiangyue Jin, Xianzu Wang, Xiao Bi, Xiaodong Liu, Xiaohan Wang, Xiaojin Shen, Xiaokang Chen, Xiaokang Zhang, Xiaosha Chen, Xiaotao Nie, Xiaowen Sun, Xiaoxiang Wang, Xin Cheng, Xin Liu, Xin Xie, Xingchao Liu, Xingkai Yu, Xinnan Song, Xinxia Shan, Xinyi Zhou, Xinyu Yang, Xinyuan Li, Xuecheng Su, Xuheng Lin, Y.~K. Li, Y.~Q. Wang, Y.~X. Wei, Y.~X. Zhu, Yang Zhang, Yanhong Xu, Yanhong Xu, Yanping Huang, Yao Li, Yao Zhao, Yaofeng Sun, Yaohui Li, Yaohui Wang, Yi~Yu, Yi~Zheng, Yichao Zhang, Yifan Shi, Yiliang Xiong, Ying He, Ying Tang, Yishi Piao, Yisong Wang, Yixuan Tan, Yiyang Ma, Yiyuan Liu, Yongqiang Guo, Yu~Wu, Yuan Ou, Yuchen Zhu, Yuduan Wang, Yue Gong, Yuheng Zou, Yujia He, Yukun Zha, Yunfan Xiong, Yunxian Ma, Yuting Yan, Yuxiang
  Luo, Yuxiang You, Yuxuan Liu, Yuyang Zhou, Z.~F. Wu, Z.~Z. Ren, Zehui Ren, Zhangli Sha, Zhe Fu, Zhean Xu, Zhen Huang, Zhen Zhang, Zhenda Xie, Zhengyan Zhang, Zhewen Hao, Zhibin Gou, Zhicheng Ma, Zhigang Yan, Zhihong Shao, Zhipeng Xu, Zhiyu Wu, Zhongyu Zhang, Zhuoshu Li, Zihui Gu, Zijia Zhu, Zijun Liu, Zilin Li, Ziwei Xie, Ziyang Song, Ziyi Gao, and Zizheng Pan. 2024.
\newblock \href {https://arxiv.org/abs/2412.19437} {Deepseek-v3 technical report}.

\bibitem[{Fadeeva et~al.(2023)Fadeeva, Vashurin, Tsvigun, Vazhentsev, Petrakov, Fedyanin, Vasilev, Goncharova, Panchenko, Panov, Baldwin, and Shelmanov}]{fadeeva-etal-2023-lm}
Ekaterina Fadeeva, Roman Vashurin, Akim Tsvigun, Artem Vazhentsev, Sergey Petrakov, Kirill Fedyanin, Daniil Vasilev, Elizaveta Goncharova, Alexander Panchenko, Maxim Panov, Timothy Baldwin, and Artem Shelmanov. 2023.
\newblock \href {https://doi.org/10.18653/v1/2023.emnlp-demo.41} {{LM}-polygraph: Uncertainty estimation for language models}.
\newblock In \emph{Proceedings of the 2023 Conference on Empirical Methods in Natural Language Processing: System Demonstrations}, pages 446--461, Singapore. Association for Computational Linguistics.

\bibitem[{Gekhman et~al.(2024)Gekhman, Yona, Aharoni, Eyal, Feder, Reichart, and Herzig}]{gekhman-etal-2024-fine}
Zorik Gekhman, Gal Yona, Roee Aharoni, Matan Eyal, Amir Feder, Roi Reichart, and Jonathan Herzig. 2024.
\newblock \href {https://doi.org/10.18653/v1/2024.emnlp-main.444} {Does fine-tuning {LLM}s on new knowledge encourage hallucinations?}
\newblock In \emph{Proceedings of the 2024 Conference on Empirical Methods in Natural Language Processing}, pages 7765--7784, Miami, Florida, USA. Association for Computational Linguistics.

\bibitem[{Han et~al.(2024)Han, Li, Chen, Shi, Du, Xiao, Liang, and Lin}]{han2024enhancingconfidenceexpressionlarge}
Haixia Han, Tingyun Li, Shisong Chen, Jie Shi, Chengyu Du, Yanghua Xiao, Jiaqing Liang, and Xin Lin. 2024.
\newblock \href {https://arxiv.org/abs/2404.10315} {Enhancing confidence expression in large language models through learning from past experience}.

\bibitem[{Hu et~al.(2022)Hu, Shen, Wallis, Allen{-}Zhu, Li, Wang, Wang, and Chen}]{Hu2021LoRALA}
Edward~J. Hu, Yelong Shen, Phillip Wallis, Zeyuan Allen{-}Zhu, Yuanzhi Li, Shean Wang, Lu~Wang, and Weizhu Chen. 2022.
\newblock \href {https://openreview.net/forum?id=nZeVKeeFYf9} {Lora: Low-rank adaptation of large language models}.
\newblock In \emph{The Tenth International Conference on Learning Representations, {ICLR} 2022, Virtual Event, April 25-29, 2022}. OpenReview.net.

\bibitem[{Huang et~al.(2023)Huang, Yu, Ma, Zhong, Feng, Wang, Chen, Peng, Feng, Qin, and Liu}]{huang2023survey}
Lei Huang, Weijiang Yu, Weitao Ma, Weihong Zhong, Zhangyin Feng, Haotian Wang, Qianglong Chen, Weihua Peng, Xiaocheng Feng, Bing Qin, and Ting Liu. 2023.
\newblock \href {https://arxiv.org/abs/2311.05232} {A survey on hallucination in large language models: Principles, taxonomy, challenges, and open questions}.
\newblock \emph{Preprint}, arXiv:2311.05232.

\bibitem[{Huang et~al.(2024)Huang, Liu, Thirukovalluru, Cohan, and Dhingra}]{huang-etal-2024-calibrating}
Yukun Huang, Yixin Liu, Raghuveer Thirukovalluru, Arman Cohan, and Bhuwan Dhingra. 2024.
\newblock \href {https://doi.org/10.18653/v1/2024.findings-emnlp.785} {Calibrating long-form generations from large language models}.
\newblock In \emph{Findings of the Association for Computational Linguistics: EMNLP 2024}, pages 13441--13460, Miami, Florida, USA. Association for Computational Linguistics.

\bibitem[{Jiang et~al.(2023)Jiang, Sablayrolles, Mensch, Bamford, Chaplot, de~las Casas, Bressand, Lengyel, Lample, Saulnier, Lavaud, Lachaux, Stock, Scao, Lavril, Wang, Lacroix, and Sayed}]{jiang2023mistral}
Albert~Q. Jiang, Alexandre Sablayrolles, Arthur Mensch, Chris Bamford, Devendra~Singh Chaplot, Diego de~las Casas, Florian Bressand, Gianna Lengyel, Guillaume Lample, Lucile Saulnier, Lélio~Renard Lavaud, Marie-Anne Lachaux, Pierre Stock, Teven~Le Scao, Thibaut Lavril, Thomas Wang, Timothée Lacroix, and William~El Sayed. 2023.
\newblock \href {https://arxiv.org/abs/2310.06825} {Mistral 7{B}}.
\newblock \emph{Preprint}, arXiv:2310.06825.

\bibitem[{Jiang et~al.(2024)Jiang, Ruan, Sattigeri, Roukos, and Hashimoto}]{jiang2024graphbased}
Mingjian Jiang, Yangjun Ruan, Prasanna Sattigeri, Salim Roukos, and Tatsunori Hashimoto. 2024.
\newblock \href {https://arxiv.org/abs/2410.20783} {Graph-based uncertainty metrics for long-form language model outputs}.
\newblock \emph{Preprint}, arXiv:2410.20783.

\bibitem[{Joshi et~al.(2017{\natexlab{a}})Joshi, Choi, Weld, and Zettlemoyer}]{joshi2017triviaqalargescaledistantly}
Mandar Joshi, Eunsol Choi, Daniel Weld, and Luke Zettlemoyer. 2017{\natexlab{a}}.
\newblock \href {https://doi.org/10.18653/v1/P17-1147} {{T}rivia{QA}: A large scale distantly supervised challenge dataset for reading comprehension}.
\newblock In \emph{Proceedings of the 55th Annual Meeting of the Association for Computational Linguistics (Volume 1: Long Papers)}, pages 1601--1611, Vancouver, Canada. Association for Computational Linguistics.

\bibitem[{Joshi et~al.(2017{\natexlab{b}})Joshi, Choi, Weld, and Zettlemoyer}]{joshi-etal-2017-triviaqa}
Mandar Joshi, Eunsol Choi, Daniel Weld, and Luke Zettlemoyer. 2017{\natexlab{b}}.
\newblock \href {https://doi.org/10.18653/v1/P17-1147} {{T}rivia{QA}: A large scale distantly supervised challenge dataset for reading comprehension}.
\newblock In \emph{Proceedings of the 55th Annual Meeting of the Association for Computational Linguistics (Volume 1: Long Papers)}, pages 1601--1611, Vancouver, Canada. Association for Computational Linguistics.

\bibitem[{Kim et~al.(2024)Kim, Liao, Vorvoreanu, Ballard, and Vaughan}]{kim2024m}
Sunnie~SY Kim, Q~Vera Liao, Mihaela Vorvoreanu, Stephanie Ballard, and Jennifer~Wortman Vaughan. 2024.
\newblock " i'm not sure, but...": Examining the impact of large language models' uncertainty expression on user reliance and trust.
\newblock In \emph{Proceedings of the 2024 ACM Conference on Fairness, Accountability, and Transparency}, pages 822--835.

\bibitem[{Kingma and Ba(2015)}]{kingma2014adam}
Diederik~P. Kingma and Jimmy Ba. 2015.
\newblock \href {http://arxiv.org/abs/1412.6980} {Adam: {A} method for stochastic optimization}.
\newblock In \emph{3rd International Conference on Learning Representations, {ICLR} 2015, San Diego, CA, USA, May 7-9, 2015, Conference Track Proceedings}.

\bibitem[{Kuhn et~al.(2023)Kuhn, Gal, and Farquhar}]{kuhn2022semantic}
Lorenz Kuhn, Yarin Gal, and Sebastian Farquhar. 2023.
\newblock \href {https://openreview.net/pdf?id=VD-AYtP0dve} {Semantic uncertainty: Linguistic invariances for uncertainty estimation in natural language generation}.
\newblock In \emph{The Eleventh International Conference on Learning Representations, {ICLR} 2023, Kigali, Rwanda, May 1-5, 2023}. OpenReview.net.

\bibitem[{Kwiatkowski et~al.(2019)Kwiatkowski, Palomaki, Redfield, Collins, Parikh, Alberti, Epstein, Polosukhin, Devlin, Lee, Toutanova, Jones, Kelcey, Chang, Dai, Uszkoreit, Le, and Petrov}]{kwiatkowski-etal-2019-natural}
Tom Kwiatkowski, Jennimaria Palomaki, Olivia Redfield, Michael Collins, Ankur Parikh, Chris Alberti, Danielle Epstein, Illia Polosukhin, Jacob Devlin, Kenton Lee, Kristina Toutanova, Llion Jones, Matthew Kelcey, Ming-Wei Chang, Andrew~M. Dai, Jakob Uszkoreit, Quoc Le, and Slav Petrov. 2019.
\newblock \href {https://doi.org/10.1162/tacl_a_00276} {Natural questions: A benchmark for question answering research}.
\newblock \emph{Transactions of the Association for Computational Linguistics}, 7:452--466.

\bibitem[{Kwon et~al.(2023)Kwon, Li, Zhuang, Sheng, Zheng, Yu, Gonzalez, Zhang, and Stoica}]{kwon2023efficient}
Woosuk Kwon, Zhuohan Li, Siyuan Zhuang, Ying Sheng, Lianmin Zheng, Cody~Hao Yu, Joseph Gonzalez, Hao Zhang, and Ion Stoica. 2023.
\newblock Efficient memory management for large language model serving with pagedattention.
\newblock In \emph{Proceedings of the 29th Symposium on Operating Systems Principles}, pages 611--626.

\bibitem[{Li et~al.(2024)Li, Tang, and Yang}]{li2024know}
Jiaqi Li, Yixuan Tang, and Yi~Yang. 2024.
\newblock \href {https://arxiv.org/abs/2406.10099} {Know the unknown: An uncertainty-sensitive method for llm instruction tuning}.

\bibitem[{Li et~al.(2023)Li, Cheng, Zhao, Nie, and Wen}]{li2023haluevallargescalehallucinationevaluation}
Junyi Li, Xiaoxue Cheng, Xin Zhao, Jian-Yun Nie, and Ji-Rong Wen. 2023.
\newblock \href {https://doi.org/10.18653/v1/2023.emnlp-main.397} {{H}alu{E}val: A large-scale hallucination evaluation benchmark for large language models}.
\newblock In \emph{Proceedings of the 2023 Conference on Empirical Methods in Natural Language Processing}, pages 6449--6464, Singapore. Association for Computational Linguistics.

\bibitem[{Lin et~al.(2022)Lin, Hilton, and Evans}]{lin2022teaching}
Stephanie Lin, Jacob Hilton, and Owain Evans. 2022.
\newblock \href {https://arxiv.org/abs/2205.14334} {Teaching models to express their uncertainty in words}.
\newblock \emph{ArXiv preprint}, abs/2205.14334.

\bibitem[{Lin et~al.(2023)Lin, Trivedi, and Sun}]{lin2023generating}
Zhen Lin, Shubhendu Trivedi, and Jimeng Sun. 2023.
\newblock \href {https://arxiv.org/abs/2305.19187} {Generating with confidence: Uncertainty quantification for black-box large language models}.
\newblock \emph{Preprint}, arXiv:2305.19187.

\bibitem[{Liu and Wu(2024)}]{liu2024multi}
Terrance Liu and Zhiwei~Steven Wu. 2024.
\newblock \href {https://arxiv.org/abs/2407.21057} {Multi-group uncertainty quantification for long-form text generation}.

\bibitem[{Madaan et~al.(2023)Madaan, Tandon, Gupta, Hallinan, Gao, Wiegreffe, Alon, Dziri, Prabhumoye, Yang, Gupta, Majumder, Hermann, Welleck, Yazdanbakhsh, and Clark}]{madaan2023self}
Aman Madaan, Niket Tandon, Prakhar Gupta, Skyler Hallinan, Luyu Gao, Sarah Wiegreffe, Uri Alon, Nouha Dziri, Shrimai Prabhumoye, Yiming Yang, Shashank Gupta, Bodhisattwa~Prasad Majumder, Katherine Hermann, Sean Welleck, Amir Yazdanbakhsh, and Peter Clark. 2023.
\newblock \href {http://papers.nips.cc/paper\_files/paper/2023/hash/91edff07232fb1b55a505a9e9f6c0ff3-Abstract-Conference.html} {Self-refine: Iterative refinement with self-feedback}.
\newblock In \emph{Advances in Neural Information Processing Systems 36: Annual Conference on Neural Information Processing Systems 2023, NeurIPS 2023, New Orleans, LA, USA, December 10 - 16, 2023}.

\bibitem[{Meta(2024)}]{llama3modelcard}
Meta. 2024.
\newblock \href {https://github.com/meta-llama/llama3/blob/main/MODEL_CARD.md} {Llama 3 model card}.

\bibitem[{Min et~al.(2023{\natexlab{a}})Min, Krishna, Lyu, Lewis, Yih, Koh, Iyyer, Zettlemoyer, and Hajishirzi}]{min-etal-2023-factscore}
Sewon Min, Kalpesh Krishna, Xinxi Lyu, Mike Lewis, Wen-tau Yih, Pang Koh, Mohit Iyyer, Luke Zettlemoyer, and Hannaneh Hajishirzi. 2023{\natexlab{a}}.
\newblock \href {https://doi.org/10.18653/v1/2023.emnlp-main.741} {{FA}ct{S}core: Fine-grained atomic evaluation of factual precision in long form text generation}.
\newblock In \emph{Proceedings of the 2023 Conference on Empirical Methods in Natural Language Processing}, pages 12076--12100, Singapore. Association for Computational Linguistics.

\bibitem[{Min et~al.(2023{\natexlab{b}})Min, Krishna, Lyu, Lewis, Yih, Koh, Iyyer, Zettlemoyer, and Hajishirzi}]{min2023factscorefinegrainedatomicevaluation}
Sewon Min, Kalpesh Krishna, Xinxi Lyu, Mike Lewis, Wen-tau Yih, Pang Koh, Mohit Iyyer, Luke Zettlemoyer, and Hannaneh Hajishirzi. 2023{\natexlab{b}}.
\newblock \href {https://doi.org/10.18653/v1/2023.emnlp-main.741} {{FA}ct{S}core: Fine-grained atomic evaluation of factual precision in long form text generation}.
\newblock In \emph{Proceedings of the 2023 Conference on Empirical Methods in Natural Language Processing}, pages 12076--12100, Singapore. Association for Computational Linguistics.

\bibitem[{OpenAI et~al.(2024)OpenAI, :, Hurst, Lerer, Goucher, Perelman, Ramesh, Clark, Ostrow, Welihinda, Hayes, Radford, Mądry, Baker-Whitcomb, Beutel, Borzunov, Carney, Chow, Kirillov et~al.}]{gpt4o}
OpenAI, :, Aaron Hurst, Adam Lerer, Adam~P. Goucher, Adam Perelman, Aditya Ramesh, Aidan Clark, AJ~Ostrow, Akila Welihinda, Alan Hayes, Alec Radford, Aleksander Mądry, Alex Baker-Whitcomb, Alex Beutel, Alex Borzunov, Alex Carney, Alex Chow, Alex Kirillov, et~al. 2024.
\newblock \href {https://arxiv.org/abs/2410.21276} {Gpt-4o system card}.
\newblock \emph{Preprint}, arXiv:2410.21276.

\bibitem[{OpenAI(2022)}]{gpt3.5}
OpenAI. 2022.
\newblock Chatgpt blog post.
\newblock \url{https://openai.com/blog/chatgpt}.
\newblock Accessed: 2024-09-06.

\bibitem[{Song et~al.(2024)Song, Kim, and Iyyer}]{song2024veriscore}
Yixiao Song, Yekyung Kim, and Mohit Iyyer. 2024.
\newblock \href {https://arxiv.org/abs/2406.19276} {Veriscore: Evaluating the factuality of verifiable claims in long-form text generation}.
\newblock \emph{Preprint}, arXiv:2406.19276.

\bibitem[{Wang et~al.(2024)Wang, Geng, Wang, Wang, Fu, and Zheng}]{wang2024sampleidentifygeneralframework}
Qingni Wang, Tiantian Geng, Zhiyuan Wang, Teng Wang, Bo~Fu, and Feng Zheng. 2024.
\newblock \href {https://arxiv.org/abs/2410.08174} {Sample then identify: A general framework for risk control and assessment in multimodal large language models}.
\newblock \emph{Preprint}, arXiv:2410.08174.

\bibitem[{Wang et~al.(2025)Wang, Wang, Zhang, Chen, Zhu, Shi, and Xu}]{wang2025sconuselectiveconformaluncertainty}
Zhiyuan Wang, Qingni Wang, Yue Zhang, Tianlong Chen, Xiaofeng Zhu, Xiaoshuang Shi, and Kaidi Xu. 2025.
\newblock \href {https://arxiv.org/abs/2504.14154} {Sconu: Selective conformal uncertainty in large language models}.
\newblock \emph{Preprint}, arXiv:2504.14154.

\bibitem[{Wei et~al.(2024{\natexlab{a}})Wei, Karina, Chung, Jiao, Papay, Glaese, Schulman, and Fedus}]{wei2024measuringshortformfactualitylarge}
Jason Wei, Nguyen Karina, Hyung~Won Chung, Yunxin~Joy Jiao, Spencer Papay, Amelia Glaese, John Schulman, and William Fedus. 2024{\natexlab{a}}.
\newblock \href {https://arxiv.org/abs/2411.04368} {Measuring short-form factuality in large language models}.

\bibitem[{Wei et~al.(2024{\natexlab{b}})Wei, Yang, Song, Lu, Hu, Huang, Tran, Peng, Liu, Huang, Du, and Le}]{wei2024longfact}
Jerry Wei, Chengrun Yang, Xinying Song, Yifeng Lu, Nathan Hu, Jie Huang, Dustin Tran, Daiyi Peng, Ruibo Liu, Da~Huang, Cosmo Du, and Quoc~V. Le. 2024{\natexlab{b}}.
\newblock \href {http://papers.nips.cc/paper\_files/paper/2024/hash/937ae0e83eb08d2cb8627fe1def8c751-Abstract-Conference.html} {Long-form factuality in large language models}.
\newblock In \emph{Advances in Neural Information Processing Systems 38: Annual Conference on Neural Information Processing Systems 2024, NeurIPS 2024, Vancouver, BC, Canada, December 10 - 15, 2024}.

\bibitem[{Wei et~al.(2024{\natexlab{c}})Wei, Yang, Song, Lu, Hu, Huang, Tran, Peng, Liu, Huang, Du, and Le}]{wei2024longformfactualitylargelanguage}
Jerry Wei, Chengrun Yang, Xinying Song, Yifeng Lu, Nathan Hu, Jie Huang, Dustin Tran, Daiyi Peng, Ruibo Liu, Da~Huang, Cosmo Du, and Quoc~V. Le. 2024{\natexlab{c}}.
\newblock \href {http://papers.nips.cc/paper\_files/paper/2024/hash/937ae0e83eb08d2cb8627fe1def8c751-Abstract-Conference.html} {Long-form factuality in large language models}.
\newblock In \emph{Advances in Neural Information Processing Systems 38: Annual Conference on Neural Information Processing Systems 2024, NeurIPS 2024, Vancouver, BC, Canada, December 10 - 15, 2024}.

\bibitem[{Xu et~al.(2024)Xu, Wu, Diao, Liu, Wang, Chen, and Gao}]{xu2024sayself}
Tianyang Xu, Shujin Wu, Shizhe Diao, Xiaoze Liu, Xingyao Wang, Yangyi Chen, and Jing Gao. 2024.
\newblock \href {https://arxiv.org/abs/2405.20974} {Sayself: Teaching llms to express confidence with self-reflective rationales}.
\newblock \emph{Preprint}, arXiv:2405.20974.

\bibitem[{Yang et~al.(2024{\natexlab{a}})Yang, Yang, Hui, Zheng, Yu, Zhou, Li, Li, Liu, Huang, Dong, Wei, Lin, Tang, Wang, Yang, Tu, Zhang, Ma, Yang, Xu, Zhou, Bai, He, Lin, Dang, Lu, Chen, Yang, Li, Xue, Ni, Zhang, Wang, Peng, Men, Gao, Lin, Wang, Bai, Tan, Zhu, Li, Liu, Ge, Deng, Zhou, Ren, Zhang, Wei, Ren, Liu, Fan, Yao, Zhang, Wan, Chu, Liu, Cui, Zhang, Guo, and Fan}]{yang2024qwen2technicalreport}
An~Yang, Baosong Yang, Binyuan Hui, Bo~Zheng, Bowen Yu, Chang Zhou, Chengpeng Li, Chengyuan Li, Dayiheng Liu, Fei Huang, Guanting Dong, Haoran Wei, Huan Lin, Jialong Tang, Jialin Wang, Jian Yang, Jianhong Tu, Jianwei Zhang, Jianxin Ma, Jianxin Yang, Jin Xu, Jingren Zhou, Jinze Bai, Jinzheng He, Junyang Lin, Kai Dang, Keming Lu, Keqin Chen, Kexin Yang, Mei Li, Mingfeng Xue, Na~Ni, Pei Zhang, Peng Wang, Ru~Peng, Rui Men, Ruize Gao, Runji Lin, Shijie Wang, Shuai Bai, Sinan Tan, Tianhang Zhu, Tianhao Li, Tianyu Liu, Wenbin Ge, Xiaodong Deng, Xiaohuan Zhou, Xingzhang Ren, Xinyu Zhang, Xipin Wei, Xuancheng Ren, Xuejing Liu, Yang Fan, Yang Yao, Yichang Zhang, Yu~Wan, Yunfei Chu, Yuqiong Liu, Zeyu Cui, Zhenru Zhang, Zhifang Guo, and Zhihao Fan. 2024{\natexlab{a}}.
\newblock \href {https://arxiv.org/abs/2407.10671} {Qwen2 technical report}.

\bibitem[{Yang et~al.(2025)Yang, Zhang, Zhang, Huang, Yang, Collier, Yu, and Yang}]{logu}
Ruihan Yang, Caiqi Zhang, Zhisong Zhang, Xinting Huang, Sen Yang, Nigel Collier, Dong Yu, and Deqing Yang. 2025.
\newblock \href {https://doi.org/10.18653/v1/2025.acl-long.928} {{L}o{GU}: Long-form generation with uncertainty expressions}.
\newblock In \emph{Proceedings of the 63rd Annual Meeting of the Association for Computational Linguistics (Volume 1: Long Papers)}, pages 18947--18968, Vienna, Austria. Association for Computational Linguistics.

\bibitem[{Yang et~al.(2024{\natexlab{b}})Yang, Chern, Qiu, Neubig, and Liu}]{yang2023alignment}
Yuqing Yang, Ethan Chern, Xipeng Qiu, Graham Neubig, and Pengfei Liu. 2024{\natexlab{b}}.
\newblock \href {http://papers.nips.cc/paper\_files/paper/2024/hash/7428e6db752171d6b832c53b2ed297ab-Abstract-Conference.html} {Alignment for honesty}.
\newblock In \emph{Advances in Neural Information Processing Systems 38: Annual Conference on Neural Information Processing Systems 2024, NeurIPS 2024, Vancouver, BC, Canada, December 10 - 15, 2024}.

\bibitem[{Zhang et~al.(2024{\natexlab{a}})Zhang, Liu, Basaldella, and Collier}]{zhang-etal-2024-luq}
Caiqi Zhang, Fangyu Liu, Marco Basaldella, and Nigel Collier. 2024{\natexlab{a}}.
\newblock \href {https://doi.org/10.18653/v1/2024.emnlp-main.299} {{LUQ}: Long-text uncertainty quantification for {LLM}s}.
\newblock In \emph{Proceedings of the 2024 Conference on Empirical Methods in Natural Language Processing}, pages 5244--5262, Miami, Florida, USA. Association for Computational Linguistics.

\bibitem[{Zhang et~al.(2024{\natexlab{b}})Zhang, Yang, Zhang, Huang, Yang, Yu, and Collier}]{zhang2024atomic}
Caiqi Zhang, Ruihan Yang, Zhisong Zhang, Xinting Huang, Sen Yang, Dong Yu, and Nigel Collier. 2024{\natexlab{b}}.
\newblock \href {https://arxiv.org/abs/2410.13246} {Atomic calibration of llms in long-form generations}.

\bibitem[{Zhang et~al.(2024{\natexlab{c}})Zhang, Diao, Lin, Fung, Lian, Wang, Chen, Ji, and Zhang}]{zhang-etal-2024-r}
Hanning Zhang, Shizhe Diao, Yong Lin, Yi~Fung, Qing Lian, Xingyao Wang, Yangyi Chen, Heng Ji, and Tong Zhang. 2024{\natexlab{c}}.
\newblock \href {https://doi.org/10.18653/v1/2024.naacl-long.394} {{R}-tuning: Instructing large language models to say {\textquoteleft}{I} don`t know'}.
\newblock In \emph{Proceedings of the 2024 Conference of the North American Chapter of the Association for Computational Linguistics: Human Language Technologies (Volume 1: Long Papers)}, pages 7113--7139, Mexico City, Mexico. Association for Computational Linguistics.

\bibitem[{Zhang et~al.(2023)Zhang, Li, Cui, Cai, Liu, Fu, Huang, Zhao, Zhang, Chen et~al.}]{zhang2023siren}
Yue Zhang, Yafu Li, Leyang Cui, Deng Cai, Lemao Liu, Tingchen Fu, Xinting Huang, Enbo Zhao, Yu~Zhang, Yulong Chen, et~al. 2023.
\newblock \href {https://arxiv.org/abs/2309.01219} {Siren's song in the ai ocean: a survey on hallucination in large language models}.
\newblock \emph{ArXiv preprint}, abs/2309.01219.

\bibitem[{Zhao et~al.(2024{\natexlab{a}})Zhao, Goyal, Chiu, Jiang, Newman, Ravichander, Chandu, Bras, Cardie, Deng, and Choi}]{zhao2024wildhallu}
Wenting Zhao, Tanya Goyal, Yu~Ying Chiu, Liwei Jiang, Benjamin Newman, Abhilasha Ravichander, Khyathi Chandu, Ronan~Le Bras, Claire Cardie, Yuntian Deng, and Yejin Choi. 2024{\natexlab{a}}.
\newblock \href {https://arxiv.org/abs/2407.17468} {Wildhallucinations: Evaluating long-form factuality in llms with real-world entity queries}.

\bibitem[{Zhao et~al.(2024{\natexlab{b}})Zhao, Goyal, Chiu, Jiang, Newman, Ravichander, Chandu, Bras, Cardie, Deng, and Choi}]{zhao2024wildhallucinationsevaluatinglongformfactuality}
Wenting Zhao, Tanya Goyal, Yu~Ying Chiu, Liwei Jiang, Benjamin Newman, Abhilasha Ravichander, Khyathi Chandu, Ronan~Le Bras, Claire Cardie, Yuntian Deng, and Yejin Choi. 2024{\natexlab{b}}.
\newblock \href {https://arxiv.org/abs/2407.17468} {Wildhallucinations: Evaluating long-form factuality in llms with real-world entity queries}.

\bibitem[{Zhou et~al.(2023)Zhou, Jurafsky, and Hashimoto}]{zhou-etal-2023-navigating}
Kaitlyn Zhou, Dan Jurafsky, and Tatsunori Hashimoto. 2023.
\newblock \href {https://doi.org/10.18653/v1/2023.emnlp-main.335} {Navigating the grey area: How expressions of uncertainty and overconfidence affect language models}.
\newblock In \emph{Proceedings of the 2023 Conference on Empirical Methods in Natural Language Processing}, pages 5506--5524, Singapore. Association for Computational Linguistics.

\end{thebibliography}

\clearpage
\newpage
\appendix
\section*{Appendix}
\label{sec:appendix}

\section{Human Annotation} 
\label{app:annotation}

\subsection{Human Annotation on \uncle Construction} 
We, the authors, conducted human verification during the dataset construction process. 
We manually reviewed all the top-ranked relations (aspects) and removed those that were (1) not suitable for short-form QA, and (2) too difficult to answer or of limited importance, such as Freebase ID and IMDb ID. 
For the manually constructed questions, we also reviewed all of them to ensure they were proper and accurate short-form questions.

\subsection{Human Annotation on Evaluation Pipeline} 
We randomly selected 100 samples for human annotation. 
The two annotators were compensated above the local minimum wage. 
Both annotators had postgraduate-level English proficiency and backgrounds in computer science. 
They agreed to contribute data for our analysis. 
The statistics for some key components are as follows: (1) Accuracy of judging long-form QA as correct, incorrect, or uncertain: agreement rate of 95\% and error rate of 5\%. (2) Accuracy of judging short-form QA as correct, incorrect, or uncertain: agreement rate of 93\% and error rate of 3\%. 
Since this task primarily involves comparing answers with the ground truth, GPT-4o performs well overall. However, the errors primarily arise in the following cases: (1) when multiple correct answers exist, GPT-4o may deem partially correct answers with unfactual information as correct; (2) for numbers with many digits, GPT-4o may incorrectly treat values within a reasonable range as correct.

\section{Instruction Prompt Examples.}
\label{sec:logu_prompt}
The instruction prompts for fact-checking the answers to key aspects in long-form and short-form generation, as discussed in \S\ref{sec:original}, are provided in Listing~\ref{listing:prompt}.
\lstset{
    backgroundcolor=\color[RGB]{245,245,244},
    breaklines=true,
    breakindent=0pt,
    basicstyle=\ttfamily\small,
    emph={Factcheck, Long-Form, Long, Short, Form, Instruction, Model, Question, Questions, Gold, Answer, Answers, Paragraph, Analysis, And},
    emphstyle={\bfseries\color{brown}}
}
\begin{lstlisting}[caption={The instruction prompts for Factchecking.},label=listing:prompt]

Short Form Factcheck Instruction: 
Your task is to evaluate whether the model answer correctly answers the given questions, based on the gold answer.  

For each question, assign one of the following veracity labels:  
- UNC: If the model answer contains uncertainty expressions or explicitly refuses to answer the question. 
- S: If the model answer is definitive and does not contradict the gold answer, this includes cases where the model answer is only partially consistent with the gold answer but does not introduce incorrect information.
- NS: If the model answer is definitive but contradicts the gold answer.

Output Format:
Your responses must follow the format demonstrated in the examples below:
### Question: What is William Waldegrave, Baron Waldegrave of North Hill's date of birth? Model Answer: North Hill was born on August 15, 1946. Gold Answer: 1946-08-15T00:00:00Z. Analysis: The model answer aligns with the gold answer. $S$
### Question: What instrument does John Estes play? Model Answer: John Estes plays piano. Gold Answer: guitar. Analysis: The model answer states that John Estes plays piano, which contradicts the gold answer. $NS$
### Question: Where was John Estes born? Model Answer: I have no information on where John Estes was born. Gold Answer: Ripley. Analysis: The paragraph states, "I have no information on where John Estes was born." indicating uncertainty. $UNC$

Now, evaluate the following model answers based on the gold answers.  

{atomic_facts_string}

Now, provide your responses in the required format:  

Long Form Factcheck Instruction:
Your task is to evaluate whether the statements in the given paragraph accurately answer the provided questions, based on the gold answer.

For each question, assign one of the following veracity labels:  
- UNC: If the statement in the paragraph about this question includes uncertainty expressions, or explicitly states a lack of knowledge.  
- S: If the statement in the paragraph about this question is definitive and does not contradict the gold answer, this includes cases where the model answer is only partially consistent with the gold answer but does not introduce incorrect information.
- NS: If the statement in the paragraph about this question is definitive but contradicts the gold answer or fails to mention the question altogether.

Output Format:
Your responses must follow the format demonstrated in the examples below:
### Question: What is William Waldegrave, Baron Waldegrave of North Hill's date of birth? Gold Answer: 1946-08-15T00:00:00Z. Analysis: The paragraph states that North Hill was born on August 15, 1946, which aligns with the gold answer. $S$
### Question: What instrument does John Estes play? Gold Answer: guitar. Analysis: The paragraph states that John Estes plays piano, which contradicts the gold answer. $NS$
### Question: Where was John Estes born? Gold Answer: Ripley.  Analysis: The paragraph states, "I have no information on where John Estes was born," indicating uncertainty. $UNC$

Now, evaluate the following paragraph and questions based on the gold answers. 

Paragraph:
{paragraph}

Questions And Gold Answers:

{qa_pairs}

Now, provide your responses following the specified format:
\end{lstlisting}

\section{Teaching Models to Express Uncertainty} 
\label{app:teaching}
\subsection{Prompt-based Methods}
\label{sec:prompt_details}
Here, we list the prompts for the prompt-based methods (\ie Zero-Shot, Few-Shot, and Paired Few-Shot) in \S\ref{sec:training}. 
\lstset{
    backgroundcolor=\color[RGB]{245,245,244},
    breaklines=true,
    breakindent=0pt,
    basicstyle=\ttfamily\small,
    emph={Zero,-,Shot,(,Bios,), LongFact, WildHallu, Few, Examples, Instruction, Paired, Question, Answer, Good, Bad, Refine, Self, Long, Short, Form},
    emphstyle={\bfseries\color{brown}}
}
\begin{lstlisting}[caption={The instruction prompts of key procedures.},label=listing:baselines]
Zero Shot(Long Form):
In a paragraph, introduce the [entity], including [A1], [A2], [A3], [A4], [A5], [A6]. You should express uncertainty for any aspect you are unsure about.

Few Shot Examples(Long Form):
Your task is to write a biography for a specific entity. You should express uncertainty for any information you are not familiar with.
Question: Tell me bio of [example_entity].
Answer: [example_answer]

Paired Few Examples(Long Form):
Your task is to write a biography for a specific entity. You should express uncertainty for any information you are not familiar with.
Question: Tell me a bio of [example_entity].
Good Answer: [example_answer]

Zero Shot (Short Form):
[Question]. You should express uncertainty for any questions you are unsure about.

Few Shot Examples (Short Form):
Your task is to answer the given question. You should express uncertainty for any information you are not familiar with.
Question: [example_question]
Good Answer: [example_answer]


Paired Few Examples(Short Form):
Your task is to answer the given question. You should express uncertainty for any information you are not familiar with.
Question: [example_question]
Good Answer: [example_good_answer]
Bad Answer: [example_short_answer]
\end{lstlisting}

\subsection{Training-based Methods}
\label{sec:training_details}

\begin{table*}[ht]
\small
    \centering
    \setlength\tabcolsep{20pt}
    \begin{tabular}{lcc}
    \toprule
    Configuration & SFT(Long/Short/Mix) & DPO(Long/Short/Mix) \\
    \midrule
    Model & Mistral-7B(Llama3-8B)-Instruct  & Mistral-7B(Llama3-8B)-Instruct \\
    Number of epochs & 3 & 3 \\
    Devices &8 NVIDIA GPUs  &8 NVIDIA GPUs\\
    Total Batch size & 32 samples & 64 samples \\
    Cutoff Length & 1024 & 1024 \\
    Optimizer & Adam~\cite{kingma2014adam} & Adam~\cite{kingma2014adam} \\
    & $(\beta_1=0.9, \beta_2=0.98, \epsilon=1 \times 10^{-8})$ & $(\beta_1=0.9, \beta_2=0.98, \epsilon=1 \times 10^{-8})$\\
    Learning rate & $5 \times 10^{-5}$ & $1 \times 10^{-5}$ \\
    Warmup Ratio & 0.1 & 0.1 \\
    LoRA Target & $\mathrm{q}_{\text{proj}}, \mathrm{v}_{\text{proj}}$ & $\mathrm{q}_{\text{proj}}, \mathrm{v}_{\text{proj}}$\\
    LoRA Parameters & $r=8, \alpha=16, \text{dropout}=0.05$ & $r=8, \alpha=16, \text{dropout}=0.05$ \\
    Training Time &1h 37m 49s (1h 33m 24s) &51m 30s (1h 5m 39s) \\
    \bottomrule
    \end{tabular}
    \caption{Fine-tuning hyper-parameters.}
    \label{tab:configuration}
\end{table*}

In our experiments, we use Llama-3-8B-Instruct~\citep{llama3modelcard} and Mistral-7B-Instruct~\citep{jiang2023mistral} as base models.

\paragraph{Training Data}
We construct three types of training data: Idk-Dataset, which helps the model learn to express uncertainty for short questions; LoGU-Dataset, which is used for long-form uncertainty expression; and Mix-Dataset, which is a proportionally mixed combination of the Idk-Dataset and LoGU-Dataset.

\begin{itemize}[leftmargin=*]
\item \textbf{Idk-Dataset}~\citep{cheng2024aiassistantsknowdont}: The Idk-Dataset is constructed based on TrivialQA~\citep{joshi-etal-2017-triviaqa}. Given a question \( Q \), the model generates a set of answers \( \{A_i\}_{i=1}^{K} \) by being prompted \( K \) times. If the accuracy of these \( K \) answers falls below the predefined threshold \( \theta \), the chosen answer \( A_{\texttt{chosen}} \) is classified as the refuted answer (e.g., "This question is beyond the scope of my knowledge, and I am not sure what the answer is"). In this case, the rejected answer \( A_{\texttt{rejected}} \) is considered incorrect. If all \( K \) answers are correct, the chosen answer is classified as correct, and the rejected answer is classified as refuted. For this setup, we use \( K = 10 \) and \( \theta = 1 \). The pair \( (Q, A_{\texttt{chosen}}) \) is then used to form the dataset \( D_{\text{Short-SFT}} \), while the triplet \( (Q, A_{\texttt{chosen}}, A_{\texttt{rejected}}) \) is used to form \( D_{\text{Short-DPO}} \).

\item \textbf{LoGU-Dataset}~\citep{logu}: The LoGU framework adopts a divide-and-conquer approach. Given a question \( Q \) and its corresponding long-form answer \( A \), the LoGU-Dataset decomposes \( A \) into atomic claims. Fact-checking is then performed to identify correct claims \( C_{\text{s}} \) and incorrect claims \( C_{\text{ns}} \). The chosen answer \( A_{\texttt{chosen}} \) is formed by merging the correct atomic claims and revised versions of the incorrect claims that express uncertainty. The rejected answer \( A_{\texttt{rejected}} \) consists of the correct atomic claims, now revised to express uncertainty, and the incorrect claims \( C_{\text{ns}} \). The pair \( (Q, A_{\texttt{chosen}}) \) is used to form \( D_{\text{Long-SFT}} \), while the triplet \( (Q, A_{\texttt{chosen}}, A_{\texttt{rejected}}) \) is used to form \( D_{\text{Long-DPO}} \). The questions used to construct the LoGU-Dataset are sourced from Bios~\citep{bishop-etal-2024-longdocfactscore}, WildHallu~\citep{zhao2024wildhallu}, and LongFact~\citep{wei2024longfact}.

\item \textbf{Mix-Dataset}: The Mix-Dataset is created by proportionally combining the Idk-Dataset and LoGU-Dataset with a mixture ratio \( \xi \) (in \S \ref{sec:training}, we set \( \xi = 0.7 \)). \( D_{\text{Mix-SFT}} \) is formed by mixing \( D_{\text{Short-SFT}} \) and \( D_{\text{Long-SFT}} \) according to the ratio \( \xi \), while \( D_{\text{Mix-DPO}} \) is formed by mixing \( D_{\text{Short-DPO}} \) and \( D_{\text{Long-DPO}} \) according to the same ratio.
\end{itemize}
Following \citet{logu} and \citet{cheng2024aiassistantsknowdont}, the Long-DPO, Short-DPO, and Mix-DPO approaches all employ a two-stage training process (\ie first SFT, followed by DPO). 
To ensure fairness, we use the same amount of training data for all three methods in both the SFT and DPO stages (\ie 40k for the SFT stage and 20k for the DPO stage).

\paragraph{Fine-tuning Details}
We run SFT and DPO experiments with 8 NVIDIA GPUs. 
We conduct experiments with the LlamaFactory code base\footnote{\url{https://github.com/hiyouga/LLaMA-Factory}}.
Building upon prior research, which highlights the MLP layer as a crucial element for embedding knowledge within the LLM transformer architecture \citep{de2021editing}, we
only fine-tune the weight matrix of the attention
layer using LoRA~\citep{Hu2021LoRALA}. 
This method allows us to adjust the model's ability to express knowledge boundaries without altering its internal knowledge structure. 
The configurations of our hyper-parameters are detailed in Table~\ref{tab:configuration}.

\paragraph{Evaluation}
We use vLLM \citep{kwon2023efficient} for LLM inference tasks with the following parameters: temperature = 0.7, top-$p$ = 0.95, and a maximum output of 1024 tokens. 
For fact-checking, we set the temperature to 0. 
GPT-4o is used as the auxiliary model to perform fact-checking. 
The total cost for fact-checking 100 generations is \$0.46.

\newcolumntype{g}{>{\columncolor{Ground!7}}c}
\newcolumntype{d}{>{\columncolor{cyan!6}}c}
\newcolumntype{f}{>{\columncolor{lime!6}}c}
\newcolumntype{v}{>{\columncolor{purple!6}}c}
\begin{table*}[t!]
\centering
\setlength\tabcolsep{11pt}
\scalebox{0.8}[0.8]{ 
\begin{tabular}{lvvddvvddvvdd}
\toprule
\multirow{3}{*}{\textbf{Category}} & \multicolumn{4}{c}{\textbf{Unc-Zero}} & \multicolumn{4}{c}{\textbf{Pair-Few}} & \multicolumn{4}{c}{\textbf{Mix-DPO}} \\
\cmidrule(lr){2-5} \cmidrule(lr){6-9} \cmidrule(lr){10-13}
& \multicolumn{2}{c}{\textbf{Long-form}} & \multicolumn{2}{c}{\textbf{Short-form}} & \multicolumn{2}{c}{\textbf{Long-form}} & \multicolumn{2}{c}{\textbf{Short-form}} & \multicolumn{2}{c}{\textbf{Long-form}} & \multicolumn{2}{c}{\textbf{Short-form}} \\
\cmidrule(lr){2-3} \cmidrule(lr){4-5} \cmidrule(lr){6-7} \cmidrule(lr){8-9} \cmidrule(lr){10-11} \cmidrule(lr){12-13}
& \multicolumn{1}{c}{\textbf{UA}} & \multicolumn{1}{c}{\textbf{EA}} & \multicolumn{1}{c}{\textbf{UA}} & \multicolumn{1}{c}{\textbf{EA}} & \multicolumn{1}{c}{\textbf{UA}} & \multicolumn{1}{c}{\textbf{EA}} & \multicolumn{1}{c}{\textbf{UA}} & \multicolumn{1}{c}{\textbf{EA}} & \multicolumn{1}{c}{\textbf{UA}} & \multicolumn{1}{c}{\textbf{EA}} & \multicolumn{1}{c}{\textbf{UA}} & \multicolumn{1}{c}{\textbf{EA}} \\
\midrule
\textbf{Bios} & 40.0 & 46.4 & 35.2 & 47.9 & 52.4 & 49.5 & 13.4 & 26.6 & 51.2 & 54.9 & 50.9 & 60.7 \\
\textbf{Companies} & 33.3 & 67.6 & 0.0 & 73.6 & 50.1 & 67.4 & 22.6 & 31.3 & 50.0 & 63.6 & 39.3 & 73.3 \\
\textbf{Movies} & 0.0 & 49.2 & 50.0 & 56.4 & 25.0 & 50.3 & 37.4 & 38.7 & 45.5 & 47.6 & 75.0 & 57.2 \\
\textbf{Diseases} & 3.5 & 73.0 & 0.0 & 84.0 & 35.7 & 64.4 & 13.4 & 26.6 & 37.5 & 70.1 & 22.4 & 71.6 \\
\textbf{Planets} & 88.9 & 25.2 & 71.4 & 27.3 & 69.7 & 28.3 & 67.5 & 66.8 & 83.0 & 38.0 & 82.0 & 50.1 \\
\midrule
\textbf{AVG.} & 41.2 & 50.7 & 42.3 & 55.8 & 51.1 & 51.0 & 39.8 & 44.0 & 59.6 & 51.1 & 55.0 & 65.0 \\
\bottomrule
\end{tabular}
}
\caption{Performance of different methods across five categories, using Llama3-8B as the fixed base model.}
\label{tab:category_reults}
\end{table*}

\section{Alignment Between Short- and Long-form Expressions across Different Model Size}
\begin{figure}[ht!]
    \centering
\includegraphics[width=0.45\textwidth]{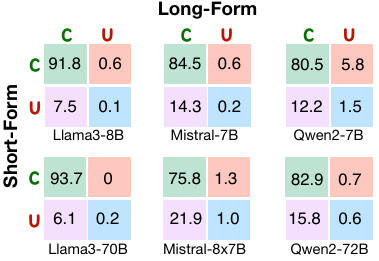}
\caption{Distribution of key aspects expressed with certainty and uncertainty by Llama3, Mistral, and Qwen2 across different model size.}
\vspace{-2mm}
\label{fig:supp_01}
\end{figure}
Using paired short- and long-form questions in \uncle, we examine whether the same aspect is consistently expressed as certain or uncertain across different QA formats, across different model size.
As shown in Figure~\ref{fig:supp_01}, both large and small models exhibit a high \colorbox{CC}{C-C} proportion. 
Larger models express more uncertainty in short-form, such as Mistral-8x7B (21.9\%+1.0\%) compared to Mistral-7B (14.3\%+0.2\%). However, in long-form, uncertainty remains low for all models, typically around 1\%. This suggests that the complexity of long-form text is a challenge for models of all sizes.

\section{Method Performance Across Categories}
\label{app:category_analysis}
We analyze the performance of different methods across five categories, using Llama3-8B as the base model.
As shown in Table~\ref{tab:category_reults}, both prompt-based and training-based methods exhibit varying performance across categories. Compared to the baseline (Unc-Zero), methods with higher overall scores (\eg Pair-Few and Mix-DPO) consistently improve results across all categories, rather than just a subset.

\section{Case Study}
\label{app:case_study}
We present a qualitative comparison between responses generated by the same base model (Llama3-8B) in two setups: a prompt-based method (Pair-Few) and a training-based method (LoGU-DPO), both answering the same question.
\definecolor{darkred}{RGB}{52,111,177}
\lstset{
    backgroundcolor=\color[RGB]{245,245,244},
    breaklines=true,
    breakindent=0pt,
    basicstyle=\ttfamily\small,
    emph={Question, Pair, Few, LoGU, DPO},
    emphstyle={\bfseries\color{brown}},
    moredelim=[is][\color{darkred}\bfseries]{@}{@} 
}
\begin{lstlisting}[caption={Comparision between generation of Prompt-based and Training-Based Methods},label=listing:case_study]
Question: In a paragraph, introduce the movie V for Vendetta, including director of photography, distributed by, film color, producer, film editor, filming location.
Pair-Few: V for Vendetta, a dystopian thriller film, was released in 2005. Directed by James McTeigue, the film is based on the graphic novel of the same name by Alan Moore and David Lloyd. The movie's director of photography was Anthony McAlpine, who captured the film's dark and gritty atmosphere. Distributed by Warner Bros. Pictures, V for Vendetta was shot in a muted color palette, @which perhaps to emphasize its themes of rebellion and resistance@. The film was produced by Joel Silver and Lana Wachowski, with Stuart Garrison serving as the film editor. The movie was primarily filmed in London, England, @though it may have used a few sets or exteriors elsewhere around the city@.
LoGU-DPO: V for Vendetta is a dystopian thriller released in 2005, directed by James McTeigue and adapted from the graphic novel by Alan Moore and David Lloyd. The film's visuals, crafted by director of photography Andrew McAlpine, employ a muted, somber color palette that underscores its themes of control and resistance. Distributed by Warner Bros. Pictures and produced by Joel Silver and the Wachowskis, the film is known for its sharp pacing and tonal consistency. @The exact identity of the film editor is unclear@. Principal photography took place in London, using a mix of iconic and understated cityscapes to portray a totalitarian future.
\end{lstlisting}
The prompt-based method tends to express uncertainty in vague or less relevant parts of the response and often makes factual errors in critical areas (\eg the identity of the film editor). 
In contrast, the training-based method handles uncertainty more appropriately, explicitly stating it when key information is unknown (\eg ``The exact identity of the film editor is unclear'').

\end{document}